\documentclass{article}

\usepackage{microtype}
\usepackage{graphicx}
\usepackage{subfigure}
\usepackage{booktabs} 
\usepackage{multirow}
\usepackage{hyperref}

\usepackage[accepted]{icml2024}

\usepackage{amsmath}
\usepackage{amssymb}
\usepackage{mathtools}
\usepackage{amsthm}

\usepackage[capitalize,noabbrev]{cleveref}

\theoremstyle{plain}

\theoremstyle{definition}

\theoremstyle{remark}

\usepackage[textsize=tiny]{todonotes}

\usepackage{amsmath,amsfonts,bm}

\def\eqref#1{equation~\ref{#1}}

\def\1{\bm{1}}

\def\vb{{\bm{b}}}

\def\vx{{\bm{x}}}
\def\vy{{\bm{y}}}

\def\mM{{\bm{M}}}

\def\mV{{\bm{V}}}
\def\mW{{\bm{W}}}

\DeclareMathAlphabet{\mathsfit}{\encodingdefault}{\sfdefault}{m}{sl}
\SetMathAlphabet{\mathsfit}{bold}{\encodingdefault}{\sfdefault}{bx}{n}

\newcommand{\R}{\mathbb{R}}

\newcommand{\sigmoid}{\sigma}

\newcommand{\normltwo}{L^2}

\icmltitlerunning{Discovering Efficient Activation Functions for Sparse LLMs}

\begin{document}

\twocolumn[
\icmltitle{ReLU$^2$ Wins: Discovering Efficient Activation Functions for Sparse LLMs}



\icmlsetsymbol{equal}{*}

\begin{icmlauthorlist}
\icmlauthor{Zhengyan Zhang}{equal,thu}
\icmlauthor{Yixin Song}{equal,sjtu}
\icmlauthor{Guanghui Yu}{ict}
\icmlauthor{Xu Han}{thu}
\icmlauthor{Yankai Lin}{ruc}
\icmlauthor{Chaojun Xiao}{thu}
\icmlauthor{Chenyang Song}{thu}
\icmlauthor{Zhiyuan Liu}{thu}
\icmlauthor{Zeyu Mi}{sjtu}
\icmlauthor{Maosong Sun}{thu}
\end{icmlauthorlist}

\icmlaffiliation{thu}{NLP Group, DCST, IAI, BNRIST, Tsinghua University, Beijing, China}
\icmlaffiliation{sjtu}{Institute of Parallel and Distributed Systems (IPADS), Shanghai Jiao Tong University, China}
\icmlaffiliation{ict}{Institute of Computing Technology, Chinese Academy of Sciences, Beijing, China}
\icmlaffiliation{ruc}{Gaoling School of Artificial Intelligence, Renmin University of China, Beijing, China}

\icmlcorrespondingauthor{Zhengyan Zhang}{zy-z19@mails.tsinghua.edu.cn}
\icmlcorrespondingauthor{Xu Han}{hanxu2022@tsinghua.edu.cn}
\icmlcorrespondingauthor{Zhiyuan Liu}{liuzy@tsinghua.edu.cn}

\icmlkeywords{Machine Learning, ICML}

\vskip 0.3in
]



\printAffiliationsAndNotice{\icmlEqualContribution} 

\begin{abstract}
Sparse computation offers a compelling solution for the inference of Large Language Models~(LLMs) in low-resource scenarios by dynamically skipping the computation of inactive neurons.
While traditional approaches focus on ReLU-based LLMs, leveraging zeros in activation values, we broaden the scope of sparse LLMs beyond zero activation values.
We introduce a general method that defines neuron activation through neuron output magnitudes and a tailored magnitude threshold, demonstrating that non-ReLU LLMs also exhibit sparse activation.
To find the most efficient activation function for sparse computation, we propose a systematic framework to examine the sparsity of LLMs from three aspects: the trade-off between sparsity and performance, the predictivity of sparsity, and the hardware affinity.
We conduct thorough experiments on LLMs utilizing different activation functions, including ReLU, SwiGLU, ReGLU, and ReLU$^2$. 
The results indicate that models employing ReLU$^2$ excel across all three evaluation aspects, highlighting its potential as an efficient activation function for sparse LLMs.
We will release the code to facilitate future research.
\end{abstract}

\section{Introduction}
Large Language Models (LLMs)~\cite{GPT-3,InstructGPT,GPT-4} have become a new paradigm in deep learning, showing a promising route to general artificial intelligence~\cite{AGI-Sparks}.
However, driving LLMs requires substantial computing and storage resources, making these models challenging to be deployed in low-resource scenarios.
A promising direction to address the issue of significant resource consumption is sparse computation, which is enabled by the observation of sparse activation in LLMs~\cite{li2023the,dejavu}.

Sparse activation refers to the phenomenon where certain model parameters contribute weakly for a given input, implying that excluding these parameters would have a negligible impact on the final model result. 
These weakly-contributed parameters are regarded as not activated for the given input, and can be skipped during inference to achieve sparse computation.
Previous efforts primarily focus on sparsely deploying the LLMs using the ReLU activation function, by utilizing the occurrence of zeros in activation values~\cite{zhang2022moefication,mirzadeh2023relu}, and have achieved promising results.
Parameters corresponding to the zero activation values contribute nothing to the final result, allowing for their exclusion in computation.

In this paper, we broaden the scope of sparse activation in LLMs by moving beyond zero activation values to include more activation functions and achieve higher sparsity ratios.
More specifically, we propose to define the neuron activation based on whether neuron output magnitudes\footnote{The output of a feed-forward network in LLMs can be treated as the summation of the outputs of all neurons in the network.} exceed a magnitude threshold, which is a more general activation definition.
Our pilot investigation studies the distribution of neuron output magnitudes within LLMs, revealing a long-tail pattern where many neuron output magnitudes are small.
In some cases, even though the activation values of neurons are not exactly zero, their output magnitudes are still small enough to be negligible.
This insight opens new perspectives, particularly evident in our discovery that a series of LLMs that do not employ ReLU, i.e., LLaMA~\cite{LLAMA2}, are also sparsely activated.
Furthermore, our analysis shows that within the LLMs using ReLU, the activation sparsity ratio formalized with long-tailed neuron output magnitudes is significantly higher than the ratio formalized with zero activation values.

As we find sparse activation is prevalent in LLMs across various activation functions, it prompts an essential question: \textit{which activation function is optimal for sparse LLMs?} Although previous works on activation function selection have focused on the performance of LLMs, we argue that the efficiency of sparse computation should also be considered so that the LLMs can proceed with efficient inference while preserving performance.
To answer this question, we consider three critical aspects: the relation between sparsity and performance, the sparsity predictivity, and the hardware affinity.
(1)~By adjusting the magnitude threshold, we explore the trade-off between sparsity and performance.
In practice, a slight performance compromise is acceptable for achieving a significantly higher sparsity ratio.
(2)~Predictivity is defined as the ability to identify inactive neurons prior to computation, which is an important factor that enables sparse activation to optimize deployment~\cite{dejavu}.
We study the predictivity of different activation functions by constructing a lightweight predictor to predict neuron activation behaviors.
(3)~Furthermore, it is crucial to acknowledge that theoretical efficiency gains do not always align with practical hardware performance.
Therefore, our investigation emphasizes how sparse computation can be optimized according to hardware features, with a particular focus on the computational relations both between tokens and between neurons.
These relations can be used to enhance memory reuse and facilitate contiguous memory access, which is pivotal for improving actual efficiency.

We conduct comprehensive experiments and analysis following the above three aspects on the LLMs with different activation functions\footnote{All model checkpoints we trained are available at \url{https://huggingface.co/SparseLLM}}, including ReLU~\cite{ReLU}, SwiGLU~\cite{SiLU}, ReGLU~\cite{GLU}, and ReLU$^2$~\cite{relu2}.
Our findings are summarized as follows:
(1)~With the same pre-training recipe and model size, models using ReGLU and ReLU$^2$ achieve comparable performance to models using SwiGLU, even though SwiGLU is the most popular activation function used by existing LLMs.
Models using ReLU$^2$ achieve the best trade-off between performance and sparsity, e.g., performance degradation is less than $0.1\%$ at a sparsity ratio close to $90\%$.
(2)~Models using ReLU and ReLU$^2$ have the highest neuron activation predictivity, which is beneficial for sparse computation in practice.
(3)~The hardware affinity of the models using ReLU and ReLU$^2$ is better than that of the models using SwiGLU and ReGLU.
Specifically, the LLMs using ReLU and ReLU$^2$ tend to activate similar neurons to process consecutive tokens and have higher frequency differences between highly co-activated neurons and lowly co-activated neurons.
As a result, the I/O overhead of feed-forward networks using ReLU$^2$ can be reduced by $92\%$ by utilizing these features.

Our contributions are summarized below:

\begin{itemize}
    \vspace{-0.5em}
    \setlength\parskip{0em}
    \setlength\parsep{0em}
    \item We reformulate a more general activation definition based on neuron output magnitudes with a magnitude threshold instead of focusing on zero activation values.
    \item We propose a systematic framework to examine sparse computation from three aspects: the trade-off between sparsity and performance, the predictivity, and the hardware affinity, to find the most efficient activation function to achieve sparse LLMs.
    \item Our experiments show ReLU$^2$ achieves good results in all three aspects, suggesting that ReLU$^2$ is a promising activation function for sparse LLMs.
\end{itemize}

\section{Related Work}
Here we mainly introduce works on deploying LLMs in low-resource scenarios. More details on LLMs can refer to the surveys~\cite {bommasani2021opportunities,zhao2023survey}.

\textbf{Efficient Inference of LLMs.}
LLM inference represents a complex challenge that necessitates a synergistic combination of algorithms and systems.
From an algorithmic perspective, researchers have explored various methods to reduce time and memory overheads, including compressing model parameters~\cite{DBLP:conf/nips/MichelLN19,DBLP:conf/nips/YaoAZWLH22,GPTQ,AWQ,DBLP:journals/corr/abs-2309-05516,DBLP:conf/icml/XiaoLSWDH23}, modifying model structures~\cite{GQA,DBLP:conf/emnlp/PengAAAABCCCDDG23,DBLP:journals/corr/abs-2312-00752,DBLP:journals/corr/abs-2304-02207,DBLP:conf/nips/ChenDWSRR21,DBLP:conf/iclr/KitaevKL20}, and optimizing decoding methods~\cite{DBLP:conf/icml/LeviathanKM23,DBLP:journals/corr/abs-2302-01318,DBLP:journals/corr/abs-2308-04623,cai2024medusa}.
On the system side, there is a focus on how computing LLMs can be seamlessly integrated with the hardware features~\cite{DBLP:conf/sc/AminabadiRALLZRSZRH22,DBLP:conf/icml/RajbhandariLYZA22,DBLP:conf/mlsys/ChenMFGTS20,DBLP:conf/ppopp/FangYZZ21,DBLP:journals/corr/abs-2211-05102,DBLP:conf/naacl/WangXWWL21,DBLP:conf/osdi/YuJKKC22}, leading to the development of more efficient frameworks like FlashAttention~\cite{DBLP:conf/nips/DaoFERR22} and vLLM~\cite{vLLM}.
Sparse activation, in particular, becomes a research area that demands an even tighter integration of algorithmic and systemic approaches.
On the one hand, the selection of activation functions and the activation predictor construction are algorithm problems.
On the other hand, how to fully exploit the sparse activation of LLMs on specific hardware is a system problem.
Researchers have achieved promising results in building efficient LLM inference systems by leveraging sparse activation~\cite{dejavu,powerinfer,flash}.

\textbf{Sparse Activation of LLMs.}
Sparse activation is a unique model property, which is widely observed in ReLU-based LLMs~\cite{zhang2022moefication,dejavu}, from T5~\cite{T5} to OPT~\cite{OPT}.
Meanwhile, this phenomenon is also observed in ReLU-based large-scale Transformers of other fields, such as Vision Transformer~\cite{li2023the}.
It has been theoretically shown that for the feed-forward network using the ReLU activation function, the activation values will be optimized to be smaller during the training process and produce sparse activation~\cite{li2023the}.
Recently, \citet{mirzadeh2023relu} has further shown that ReLU-based LLMs can achieve similar performance compared to LLMs using other mainstream activation functions while providing the opportunity to efficiently deploy LLMs through sparse computation.
However, it is still unclear whether sparse activation is unique to ReLU-based LLMs or can be observed in models using other activation functions.
Besides, there are some variants of ReLU, such as ReGLU~\cite{GLU} and ReLU$^2$~\cite{relu2}, and it is unclear the difference between these variants on the sparse features.
In this work, we explore the sparse features of LLMs using different activation functions, including ReLU variants and non-ReLU activation functions, to find the most efficient activation function for building sparse LLMs.

\textbf{Conditional Computation of LLMs.}
Conditional computation is considered a vital approach to address efficiency issues as the scale of deep neural networks expands~\cite{DBLP:conf/slsp/Bengio13}.
In the LLM era, the Mixture-of-Experts~(MoE) emerges as a representative technique for implementing this approach~\cite{Switch-Transformer,GShard}.
By employing MoE, it becomes feasible to substantially increase the model capacity without a corresponding rise in computational demand~\cite{DBLP:conf/nips/RiquelmePMNJPKH21,DBLP:conf/icml/DuHDTLXKZYFZFBZ22,artetxe-etal-2022-efficient}.
MoE introduces conditional computation before the training of LLMs while the natural emergence of sparse activation in LLMs presents a post-training opportunity to integrate conditional computation for efficient inference.
\citet{zhang2022moefication} illustrate that the sparsely-activated feed-forward network can be transformed into an MoE-like structure maintaining similar performance.
\citet{dejavu} further suggest that attention networks can also be computed in an MoE-like manner.
Our study focuses on the conditional computation of LLMs at the neuron level, which can be viewed as the most granular form of MoE where each neuron acts as an expert.
Then, sparse activation can be transformed into conditional computation to efficiently deploy LLMs in low-resource scenarios.

\section{Is Non-ReLU LLM Sparsely Activated?}
\label{sec:llama_sparsity}
Previous work has shown that LLMs using the ReLU activation function have the property of sparse activation~\cite{zhang2022moefication,li2023the}. 
Within ReLU-based LLMs, the activation values of most neurons are zero, which provides a potentiality for efficient inference of LLMs~\cite{dejavu,powerinfer}.
However, different LLMs use quite different activation functions, and some recent popular LLMs do not use ReLU as the activation function.
For example, LLaMA~\cite{llama1,LLAMA2} uses SiLU~\cite{SiLU} and Falcon~\cite{Falcon} uses GELU~\cite{GeLU} as their activation functions.
These activation functions are different from ReLU in that they are \textbf{not exactly zero} for negative inputs.
Hence, it is unclear whether these LLMs using non-ReLU activation functions also have the property of sparse activation. 
In this section, we study the activation patterns of non-ReLU LLMs, with LLaMA-2 as a representative, which is a very popular series of LLMs, to verify whether sparse activation also exists in non-ReLU LLMs.

\textbf{Definition of neuron.} Transformers~\cite{Transformer} have two main components: the multi-head attention networks and the feed-forward networks~(FFNs).
The activation functions are introduced in the FFNs to model the non-linearity of token-wise representations.
As shown in previous works~\cite{SkillNeuron,zhang2023emergent}, all FFNs of LLMs can be decomposed into combinations of neurons.
Each neuron has its own parameters, based on which we can compute the activation value and output representations of each neuron.
Here we give the decomposition of an FFN with SwiGLU adopted by LLaMA.
Please refer to Appendix~\ref{app:neuron_decomposition} for the decomposition of other variants of FFNs.
An FFN with SwiGLU is computed by
\begin{equation}
    \text{FFN}(x) = \mW^{out} [\sigmoid(\mW^{in} \vx) \odot (\mV^{in} \vx)],
\end{equation}
where $\mW^{in} \in \R^{d_{\text{ff}} \times d_{\text{model}}}$, $\mV^{in} \in \R^{d_{\text{ff}} \times d_{\text{model}}}$, $\mW^{out} \in \R^{d_{\text{model}} \times d_{\text{ff}}}$, $\sigmoid$ is the SiLU activation function, $\odot$ is the element-wise product, $d_{\text{model}}$ is the dimension of the input and output representations, and $d_{\text{ff}}$ is the dimension of the hidden layer.
We decompose the FFN along the dimension of $d_{\text{ff}}$ and get $d_{\text{ff}}$ neurons, the $i$-th of which is computed by
\begin{equation}
    a_i(x) = \sigmoid(\mW^{in}_{i,:} \vx), \text{n}_i(x) = [a_i(x) \odot (\mV^{in}_{i,:} \vx)] \mW^{out}_{:,i},
\end{equation}
where $\mW^{in}_{i,:}$, $\mV^{in}_{i,:}$ and $\mW^{out}_{:,i}$ are the $i$-th row and column of $\mW^{in}$, $\mV^{in}$ and $\mW^{out}$, respectively. $a_i(x)$ is the activation value of the $i$-th neuron, which is a scalar, and $\text{n}_i(x)$ is the output representation of the $i$-th neuron, which is a vector.
The output representation of the FFN is the summation of all neurons, i.e., $\text{FFN}(x) = \sum_{i=1}^{d_{\text{ff}}} \text{n}_i(x)$.

\textbf{Biological aspect of activation sparsity.} Although non-ReLU activation functions are not exactly zero for negative inputs, neuroscience studies reveal that biological neurons similarly transmit signals even when they are not active~\cite{breakspear2017dynamic,pariz2021transmission}.
Specifically, the dormant neurons send signals with low firing rates, which has little impact on the downstream neurons compared with the activated neurons.
Inspired by this, we study the \textbf{output representation} of each neuron $\text{n}_i(x)$ to see if there are neurons with very weak outputs. 
If such neurons exist, their contribution to the current forward pass should be negligible.
Note that the output representation of a neuron is related to the activation value of the neuron $a_i(x)$, but it is not exactly the same. Taking the FFN with SwiGLU as an example, if $\mV^{in}_{i,:}\vx = 0$, then the output representation of the $i$-th neuron is $0$ regardless of the activation value of the neuron $a_i(x)$.

\begin{figure*}
\centering
\includegraphics[width=0.9\linewidth]{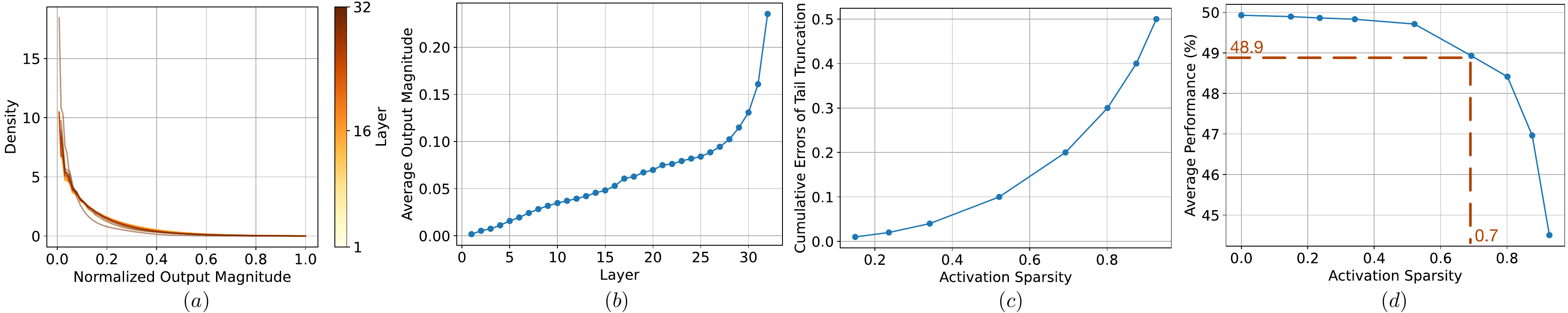}
\vspace{-0.1cm}
\caption{(a) Distribution of normalized output magnitudes of neurons in LLaMA. This distribution is long-tailed. (b) The average magnitude of neuron output representations in LLaMA with regard to the layer index. With the increase of the layer index, the average output magnitude also grows. (c) Cumulative errors of tail truncation with regard to activation sparsity. (d) Performance of LLaMA with regard to activation sparsity. The impact of activation sparsity on the performance is negligible until the sparsity ratio is larger than $0.7$.}
\label{fig:llama_7b_all}
\vspace{-0.1cm}
\end{figure*}

\textbf{Finding negligible neurons through output magnitude distribution.}
We first examine the magnitude of the output representations of neurons in LLaMA-2 7B.
If the magnitude of a neuron's output representation is extremely small, its influence in the FFN summation computations can be considered negligible.
We save the output representations of neurons in LLaMA-2 7B on a mixed corpus for analysis.
For details on the composition of the corpus, please refer to Appendix~\ref{app:experimental_details}.
Assuming the model has $L$ layers and each layer has $d_{\text{ff}}$ neurons, we record $L\times d_{\text{ff}}$ magnitudes for each token in the corpus.
The magnitude is calculated using the $\normltwo$ norm, i.e., $||\text{n}_i(x)||_2 = \sqrt{\sum_{j=1}^{d_{\text{model}}} \text{n}_{i,j}(x)^2}$.
In Figure~\ref{fig:llama_7b_all}(a) and Figure~\ref{fig:llama_7b_all}(b), we report the distribution of neuron output magnitudes for each layer.
Since the average output magnitude varies across layers, we normalize the distribution using the maximum value and report the normalized distribution in Figure~\ref{fig:llama_7b_all}(a).
We also report the average output magnitude of neurons in each layer as supplementary information in Figure~\ref{fig:llama_7b_all}(b).
From these figures, we have two observations: 
(1) The normalized distributions are quite similar across layers and show pronounced long-tail patterns.
This indicates that the output magnitudes of most neurons are small, and only a few neurons have large output magnitudes.
(2) Although the normalized distributions are similar, the average output magnitude increases progressively, consistent with observations from previous works on residual magnitudes~\cite{dejavu}. 



\textbf{Quantifying the long-tail phenomenon through cumulative errors of tail truncation.}
Based on the above observations, we aim to further quantify the impact of these long-tail neuron outputs on the FFN computation.
Here, we introduce a concept, named \textit{cumulative errors of tail truncation (CETT)}, to measure the impact of the long-tail phenomenon.
The CETT is formalized as
\begin{equation}
\small
\text{CETT}(x) = \frac{||\sum_{i \in \mathcal{D}} \text{n}_i(x)||_2}{||\text{FFN}(x)||_2}, \mathcal{D} = \{i |\, ||\text{n}_i(x)||_2 < \epsilon\},
\end{equation}
where $\epsilon$ is a threshold, $\mathcal{D}$ is the set of neurons whose output magnitudes are smaller than $\epsilon$, and $||\cdot||_2$ is the $\normltwo$ norm.
Here we pre-define several thresholds $\epsilon$ and calculate the average CETT and the corresponding sparsity ratio for each threshold.
The sparsity ratio of a single FFN for a certain token is defined as the ratio of the number of neurons whose output representations have a magnitude smaller than $\epsilon$ to the total number of neurons, i.e., $|\mathcal{D}| / d_{ff}$.
The overall sparsity ratio is the average sparsity ratio across all tokens and all FFN layers.
We report the average CETT with regard to the activation sparsity ratio on the mixed corpus in Figure~\ref{fig:llama_7b_all}(c).
From this figure, we observe that the average CETT increases much slower than the sparsity ratio.
For example, truncating about $70\%$ of the tail neurons, leading to a sparsity ratio of $0.7$, only results in a CETT of $0.2$.

We further study the task performance with regard to the sparsity ratio.
We use the datasets from Open LLM Leaderboard to evaluate the performance of LLaMA.
Details on the tasks are provided in Appendix~\ref{app:experimental_details}.
We report the average performance of LLaMA on these datasets in Figure~\ref{fig:llama_7b_all}(d).
From this figure, we observe that the performance of LLaMA is not sensitive to tail truncation when the sparsity ratio is smaller than $0.7$. Specifically, the performance only drops by about $1\%$ when the sparsity ratio is $0.7$.
However, when the sparsity ratio is larger than $0.7$, the performance drops significantly.
For example, the performance drops by over $5\%$ when the sparsity ratio reaches $0.95$.



\textbf{General definition of activation sparsity.}
Traditionally, sparse activation focuses solely on the activation value, considering neurons with an activation value of zero as inactive.
However, our findings suggest that many neurons with non-zero output in activation functions can also be deemed inactive.
Therefore, we expand the definition of activation from activation value to output magnitude.
Similar to how ReLU considers values greater than zero as active, we also need a threshold for output magnitude.
Considering the varying output magnitudes across different layers and models, this threshold needs to be adaptive.
Here we propose a threshold-finding method based on CETT because it can quantify the impact of tail neurons on the FFN computation and it is a model-agnostic metric.
Specifically, we first set a target CETT upper bound, and then find the largest output magnitude threshold that leads to a CETT smaller than the given upper bound.
In the finding process, we first calculate several quantiles of neuron output magnitudes as candidate thresholds and then find the largest candidate threshold by binary search.
We provide the pseudo-code of the threshold-finding method in Appendix~\ref{app:threshold_search}.

\section{Key Factors for Sparse LLM Deployment}

We systematically study activation functions from the perspective of sparse LLM deployment, including three key factors: the trade-off between performance and sparsity, the predictivity of the sparsity, and the hardware affinity.

(1) \textbf{Sparisty.} The sparsity ratio of LLMs forms the basis for efficiency improvement.
The sparsity ratio is defined as the ratio of the number of neurons whose magnitudes are smaller than a threshold to the total number of neurons.
A model with a higher optimization potential in deploying should have a higher sparsity ratio while not significantly sacrificing performance.
Hence we evaluate model performance across different sparsity ratios to find the best trade-off between performance and sparsity.

(2) \textbf{Predictivity.} The predictivity refers to the ability to predict the activation behaviors of neurons for a given input before the FFN computation.
If the inactive neurons can be identified before the FFN computation, the computation of these inactive neurons can be skipped to improve the inference efficiency~\cite{dejavu,powerinfer}.
With a light predictor, the efficiency gains are composed of two parts: firstly, the computational cost can be reduced; secondly, if the inference works on an offloading framework, such as requiring the transmission of parameters and intermediate results between CPU and GPU, the transmission time and the memory requirement can be further reduced.

To measure the predictivity, we adopt the predictor construction method used in previous works~\cite{dejavu,powerinfer} to build a predictor for each FFN.
The predictor is a two-layer neural network.
For more details about the predictor, please refer to Appendix~\ref{app:experimental_details}.

In the evaluation, we use two metrics: the recall of activated neurons and the ratio of the predicted inactive neurons.
The recall of activated neurons is defined as the ratio of the number of correctly predicted activated neurons to the total number of activated neurons.
The ratio of the predicted inactive neurons is defined as the ratio of the number of predicted inactive neurons to the total number of neurons.
For simplicity, we call these two metrics the \textit{recall} and the \textit{prediction sparsity}, respectively.
In the experiments, we first compute the recall and prediction sparsity per token per FFN and then compute the average recall and prediction sparsity across all tokens and FFNs.
With the same-sized predictor, a higher recall and a higher prediction sparsity indicate better predictivity of the model.

(3) \textbf{Hardware affinity.}
In practice, how to fully exploit the sparse characteristics of LLMs on specific hardware is a critical problem.
For example, when the model requires more memory than the GPU's VRAM, the model weight needs to be offloaded from the GPU's VRAM to CPU memory and flash~\cite{BMinf,Flexgen}.
The swap time of the model weight introduced by the offloading process \textbf{varies considerably} depending on the behaviors of different activation functions.
Here we focus on the computational relationships both between tokens and between neurons, which can be exploited to optimize the weight swap on real-world hardware.
These two relationships have been utilized in previous inference frameworks~\cite{flash}.


\textit{Computational relationships between tokens} aim to identify whether the activated neurons of the current token are similar to those of the previous tokens.
If there is a high overlap between the activated neurons of the current token and those of the previous tokens, a sliding window can be used to store the activated neurons of the previous tokens in a cache and the current token can reuse the parameters in the cache.
In this way, most of the parameters activated by the current token do not need to be transmitted from the storage units to the computing units, which can significantly reduce the I/O overhead of parameters.
In the experiments, we compute the average overlap between the activated neurons of the current token and those of the previous $k$ tokens by $\frac{|\mathcal{A}_i\cap(\cup_{j=1}^k\mathcal{A}_{i-j})|}{|\mathcal{A}_{i}|}$, where $\mathcal{A}_i$ is the set of activated neurons of the $i$-th token.
We denote this metric as the \textit{reuse ratio}.


\textit{Computational relationships between neurons} aim to identify whether there is a highly co-activated neuron for a given neuron than other neurons.
If so, we can directly bind the highly co-activated neuron pair to a contiguous memory address to accelerate memory access.
For example, in flash memory, the memory access time of a contiguous memory address is much shorter than that of a non-contiguous memory address when accessing the same amount of data~\cite{flash}.
In the experiments, we first calculate the co-activation frequency of each pair of neurons in a layer and then build an adjacency matrix of neurons based on the co-activation frequency, $\mM\in\mathbb{R}^{d_{\text{ff}}\times d_{\text{\textsf{ff}}}}$, where $\mM_{ij}$ is the co-activation times of the $i$-th neuron and the $j$-th neuron divided by the activation times of the $i$-th neuron and $\mM_{ii}=0$.
Since we want to compare the co-activation frequency among different models and the average co-activation frequency of different models may be different, we compute the difference between the average maximum value of each row and the average value of the whole matrix as the \textit{top-average co-activation gap}, which is formulated as $\frac{1}{d_{\text{ff}}}\sum_{i=1}^{d_{\text{ff}}}\max_{j}\mM_{ij}-\frac{1}{d_{\text{ff}}^2}\sum_{i=1}^{d_{\text{ff}}}\sum_{j=1}^{d_{\text{ff}}}\mM_{ij}$.
If the top-average co-activation gap is high, it would be beneficial to bind the highly co-activated neuron pairs to a contiguous memory address.

\section{Finding Best Function for LLM Deploying}

\begin{figure}
\centering
\includegraphics[width=\linewidth]{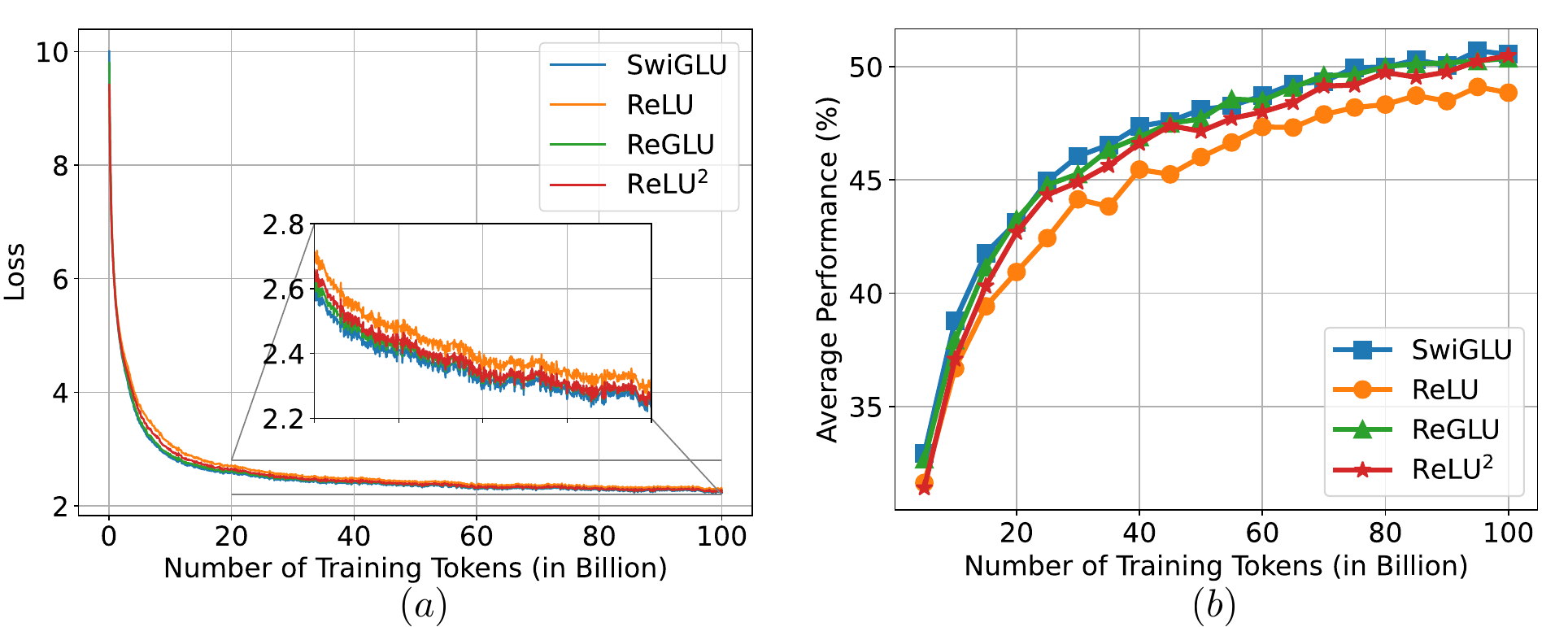}
\vspace{-0.2in}
\caption{(a) Training loss of 1B models with different activation functions. (b) Performance dynamics of 1B models on evaluation datasets. When training tokens reach 100 billion, the performance of the models with SwiGLU, ReGLU and ReLU$^2$ is very close.}
\label{fig:1b_perf}
\end{figure}

In our experiments, we analyze two series of models. 
Firstly, we train several 1.3 billion parameter models from scratch, hereafter referred to as the 1B model, utilizing different activation functions: SwiGLU, ReLU, ReGLU, and ReLU$^2$.
We strictly control the number of parameters, training data, and other hyperparameters, allowing for a direct comparison of the impact of different activation functions.
Secondly, we study larger models, specifically LLaMA-2 models of sizes from 7 billion to 70 billion parameters, as supplementary experiments to investigate the scaling behavior of the activation sparsity.
Beyond the original LLaMA-2 models using SwiGLU, we also further pre-train LLaMA models with ReGLU, which is initialized from the SwiGLU model.
This enables us to observe the trends in model sparsity as the model size increases.
The details of these models are described in Appendix~\ref{app:experimental_details}.

\begin{figure}
\centering
\includegraphics[width=\linewidth]{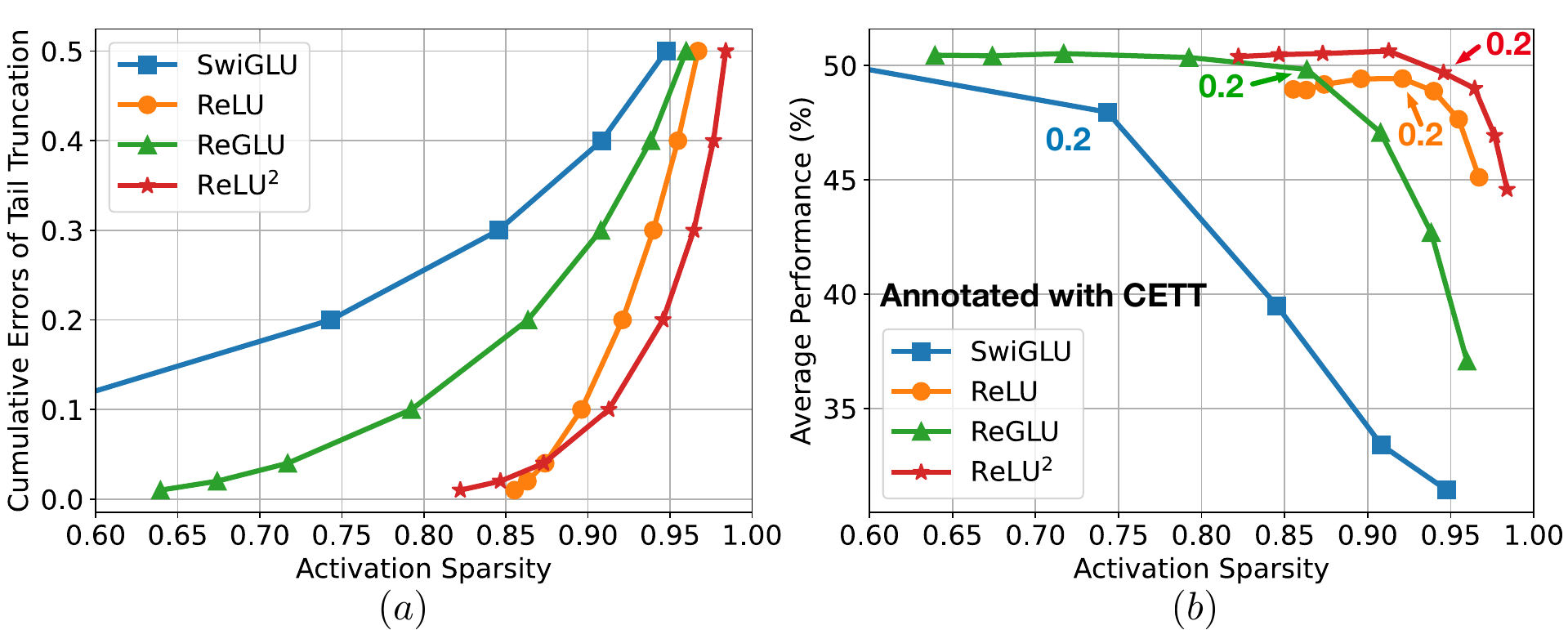}
\vspace{-0.2in}
\caption{(a) Cumulative errors of tail truncation with regard to activation sparsity. With the same cumulative errors, the sparsity of ReLU$^2$ is much higher than that of other functions in most cases. (b) Performance of 1B models under different sparsity ratios. ReLU$^2$ achieves the best trade-off between performance and sparsity. The cumulative error of $0.2$ is an inflection point of model performance for all activation functions.}
\label{fig:1b_all}
\vspace{-0.2in}
\end{figure}

We first study the training dynamics of the 1B models with different activation functions.
We report the training loss of these models in Figure~\ref{fig:1b_perf}(a) and the average performance of these models on the evaluation dataset for every 5 billion training tokens in Figure~\ref{fig:1b_perf}(b).
From these figures, we have three observations.
(1) The ReLU-based model consistently shows less optimal performance throughout the entire training process, underperforming compared to those using other activation functions.
(2) The loss and performance trajectories of SwiGLU and ReGLU are remarkably similar, suggesting a possible inherent similarity in training outcomes among different variants of GLU. This observation aligns with findings from previous research on T5 models~\cite{GLU}.
(3) The model using ReLU$^2$ initially lags slightly behind GLU models during the intermediate stages of training. However, it demonstrates a significant improvement towards the end of the training, ultimately achieving performance comparable to that of the GLU models.
In summary, ReGLU and ReLU$^2$ achieve similar performance to SwiGLU, which is widely used in existing LLMs, while ReLU consistently underperforms.
In the following experiments, we focus on the 1B models with 100 billion training tokens.

\begin{figure}
\centering
\includegraphics[width=\linewidth]{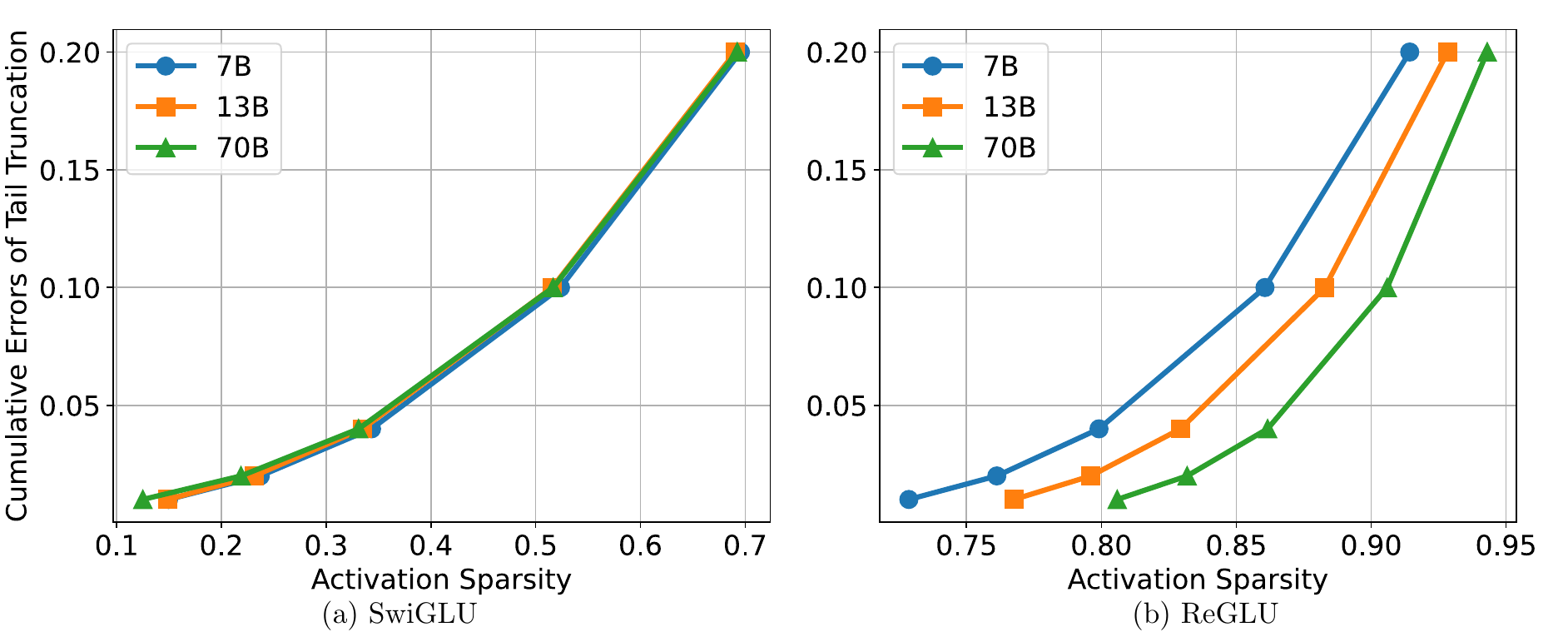}
\vspace{-0.2in}
\caption{Activation sparsity of different LLaMAs with regard to the model scale. LLaMAs with SwiGLU have a similar tendency under different model scales while LLaMAs with ReGLU become sparser with the increase of model scale.}
\label{fig:llama_sparsity_scaling}
\vspace{-0.2in}
\end{figure}

\begin{figure*}
\centering
\includegraphics[width=0.8\textwidth]{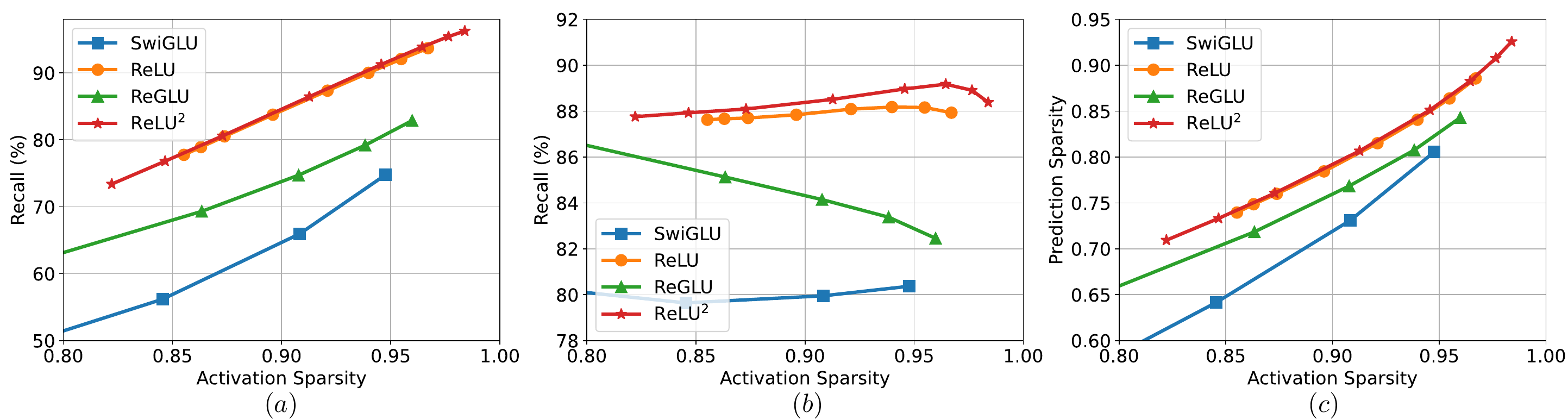}
\vspace{-0.1in}
\caption{(a) Prediction recall of 1B models with the top-$k$ prediction strategy. For each token, the neurons with the top $20\%$ largest prediction scores are predicted to be active. Hence the prediction sparsity is fixed at $0.2$. (b) Prediction recall and (c) prediction sparsity of 1B models with the threshold-based prediction strategy. The neurons with the prediction scores larger than $0.5$ are predicted to be active.
Under similar sparsity ratios, ReLU$^2$ has higher prediction recall and prediction sparsity than other activation functions.}
\label{fig:1b_pred}
\end{figure*}

\subsection{Trade-off between Performance and Sparsity}

As shown in Section~\ref{sec:llama_sparsity}, we adjust the magnitude threshold to control the sparsity ratio.
Here we study how the sparsity ratios affect the CETT of the 1B models with different activation functions and report the results in Figure~\ref{fig:1b_all}(a).
From this figure, we have two observations.
(1)~Although we have shown that the SwiGLU-based LLaMA is also sparsely activated, the sparsity ratio of the 1B model with SwiGLU is still lower than that of the models with ReLU variants.
The curve of SwiGLU is consistently above the curves of ReLU variants, indicating that with the same sparsity ratio, the CETT of SwiGLU is higher than that of ReLU variants.
(2)~Among the ReLU variants, ReLU$^2$ has the highest sparsity ratio in most cases.
When the sparsity ratio is $0.95$, the CETT of ReLU$^2$ is only $0.2$, which is nearly half of that of ReLU and ReGLU.

To further study the relationship between the sparsity and the performance, we evaluate the performance of the 1B models under different sparsity ratios on the evaluation datasets and report the results in Figure~\ref{fig:1b_all}(b).
From this figure, we have two observations.
(1) ReLU$^2$ achieves the best trade-off between performance and sparsity.
The curve of ReLU$^2$ is at the top-right corner of the figure, indicating that ReLU$^2$ has the highest sparsity ratio for a given performance and the highest performance for a given sparsity ratio.
(2) CETT is a good indicator of model performance.
The curves of different activation functions have an inflection point at the same CETT, which is $0.2$.
Before the inflection point, the performance of the models is relatively stable.
After the inflection point, the performance of the models drops significantly with the increase of the sparsity ratio.
It suggests that we can use $0.2$ as the empirical optimal CETT upper bound to search magnitude thresholds and control the sparsity ratio.

We also study the sparsity of the LLaMA models with different scales and report the results in Figure~\ref{fig:llama_sparsity_scaling}.
From this figure, we have two observations.
(1) The tendency of CETT with regard to the sparsity ratio is similar for LLaMAs with SwiGLU under different model scales.
(2) However, for LLaMAs with ReGLU, the curves of larger models are below the curves of smaller models, indicating that the impact of removing the tail neuron outputs is less significant for larger models.
It provides an opportunity to implement more efficient inference for larger LLMs using ReLU variants because the sparsity ratio increases with the model scale instead of a fixed value.


\subsection{Predictivity}

Here we study the predictivity of the 1B models with different activation functions.
Specifically, we set the predictor size to about $6\%$ of the FFN size, which has a negligible impact on the model efficiency.
We evaluate the predictor performance under two prediction strategies: the top-$k$ prediction strategy and the threshold-based prediction strategy.
For the top-$k$ prediction strategy, we predict the top $20\%$ neurons with the largest prediction scores to be active.
In this way, the prediction sparsity is fixed at $0.2$.
For the threshold-based strategy, we predict the neurons with prediction scores larger than $0.5$ to be activated.
This strategy is more flexible because we can dynamically adjust the numbers of computed neurons for different tokens, and is widely used in previous works~\cite{dejavu,powerinfer}.

We report the predictor performance under the top-$k$ prediction strategy in Figure~\ref{fig:1b_pred}(a).
From this figure, we have two observations.
(1) Under the same sparsity ratio, the prediction recalls of ReLU$^2$ and ReLU are similar and are higher than those of SwiGLU and ReGLU.
(2) The prediction recall of all activation functions increases with the sparsity ratio.
It suggests that it is easier to predict the activation behaviors of the models with higher sparsity ratios.

We report the predictor performance under the threshold-based prediction strategy in Figure~\ref{fig:1b_pred}(b) and Figure~\ref{fig:1b_pred}(c).
From these figures, we have two observations.
(1) When we remove the constraint of the prediction sparsity, the prediction sparsity of all models significantly increases with the sparsity ratio, e.g., more than $0.1$ for all models from the sparsity ratio of $0.85$ to $0.95$.
However, the corresponding prediction recalls of these models do not change much, e.g., less than $1\%$ for SwiGLU, ReLU, and $ReLU^2$ from the sparsity ratio of $0.85$ to $0.95$.
It suggests that the predictor has a higher priority to improve the prediction sparsity than the prediction recall in this strategy.
Considering that the prediction sparsity is directly related to the model efficiency, this strategy is more suitable for the scenario with limited resources.
(2) In this strategy, ReLU$^2$ still leads to the highest prediction recall and prediction sparsity among all activation functions.
It suggests that ReLU$^2$ has the highest predictivity among all activation functions.

\subsection{Hardware Affinity}

We study the hardware affinity of the 1B models from two perspectives: the computational relationships between tokens and between neurons.
These two relationships are beneficial for implementing efficient inference of LLMs on real-world hardware.
For fair comparisons, we choose specific magnitude thresholds to ensure that the sparsity ratios of all models are close to $0.95$.

\begin{figure}
\centering
\includegraphics[width=\linewidth]{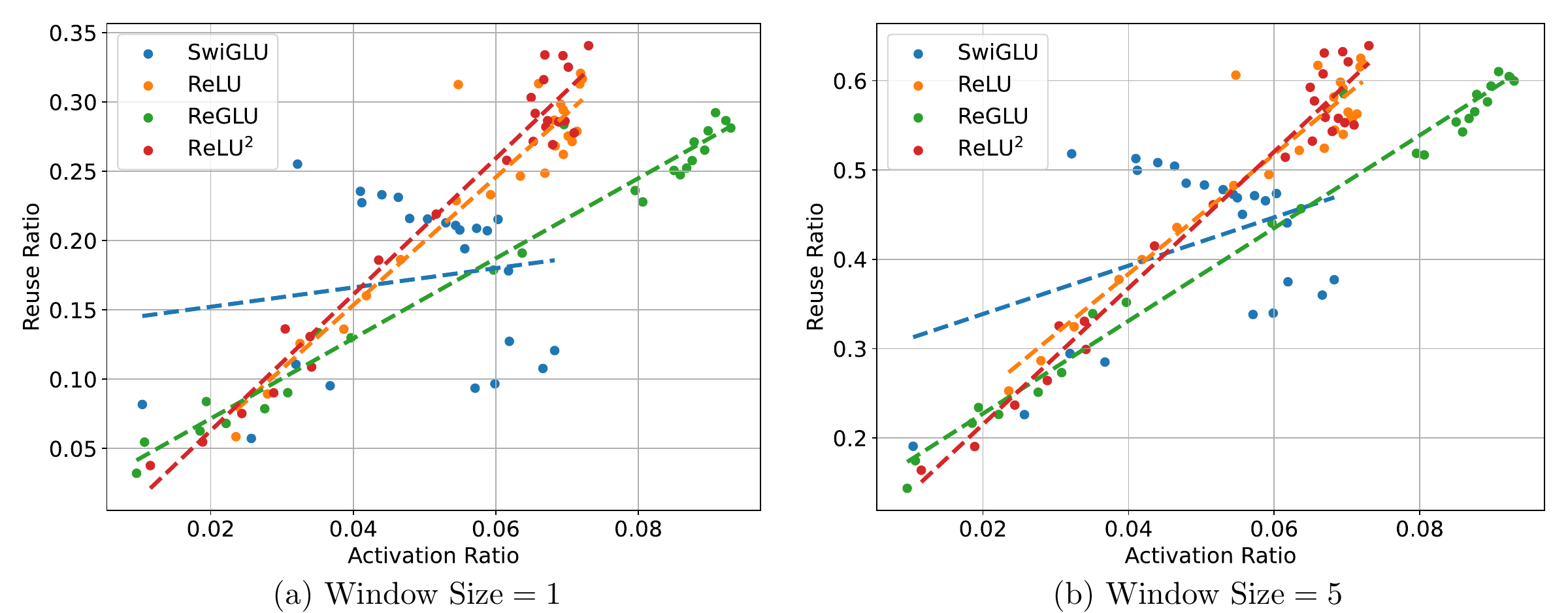}
\vspace{-0.2in}
\caption{Reuse ratios when processing consecutive tokens. Each point represents a certain layer.}
\label{fig:locality}
\vspace{-0.2in}
\end{figure}

\textbf{Computational relationships between tokens.} We report the reused neuron ratios of the 1B models with a sliding window to store the parameters of the activated neurons of the previous tokens in Figure~\ref{fig:locality}.
Each point in this figure is computed by averaging the reused neuron ratios of all tokens for a certain layer.
To supplement the reuse ratios, we also report the average activation ratios of each layer as the x-axis of this figure.
From this figure, we have three observations.
(1)~In a specific model, the activation ratios of different layers vary significantly.
For example, the activation ratios of the model with ReGLU are between $0.01$ and $0.09$ across layers.
The reuse ratios are highly correlated with the activation ratios and increase much faster than the activation ratios.
Take the model using ReLU$^2$ with window size $1$ as an example.
When the activation ratio is about $0.02$, the reuse ratio is about $0.05$.
When the activation ratio is about $0.07$, the reuse ratio is about $0.3$.
It suggests that the sliding window is more profitable for the layers with higher activation ratios.
(2)~Enlarging the sliding window can further improve the reuse ratio.
The overall reuse ratio of window size $5$ is two times that of window size $1$.
(3)~The trend of ReLU$^2$ (the red dotted line) is higher than that of other activation functions in the area with high activation ratios, indicating that ReLU$^2$ has a higher optimization potential than other activation functions.

\begin{figure}
\centering
\includegraphics[width=0.6\linewidth]{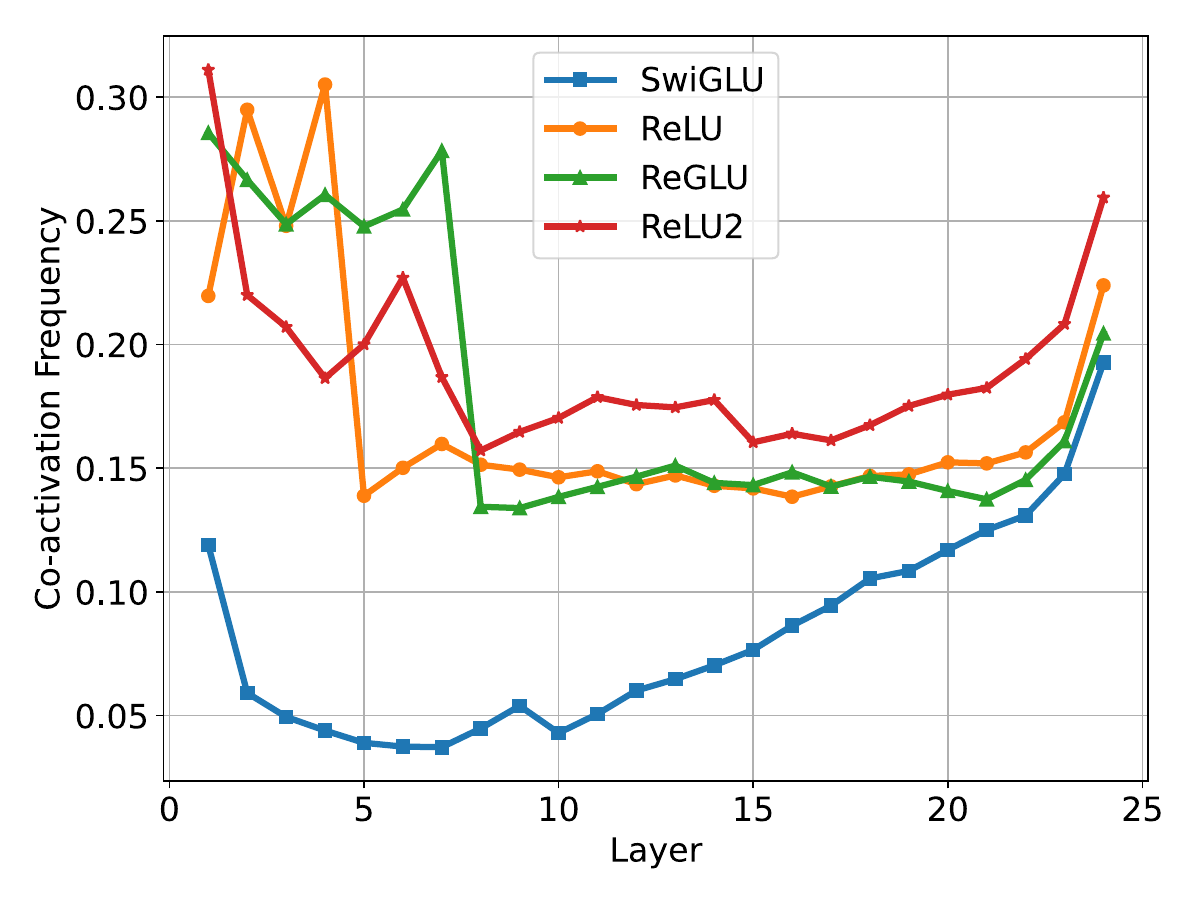}
\caption{Top-average co-activation gap of different models. The gap is defined as the difference between the top co-activation frequency and the average co-activation frequency.}
\label{fig:coact_diff}
\vspace{-0.2in}
\end{figure}

\textbf{Computational relationships between neurons.} We report the top-average co-activation gap of the 1B models in Figure~\ref{fig:coact_diff}.
From this figure, we observe that the curves of ReLU variants are similar across different layers and are above the curve of SwiGLU.
The overall top-average co-activation gap of ReLU variants is about $0.2$, which is higher than that of SwiGLU (about $0.1$).
It suggests that it is more beneficial to reallocate the memory address of highly co-activated neuron pairs to a contiguous memory address for ReLU variants than for SwiGLU.
Moreover, the top-average co-activation gap of ReLU$^2$ is consistently higher than that of ReLU and ReGLU after the eighth layer, indicating that it is more beneficial to use the reallocation strategy for ReLU$^2$ than for ReLU and ReGLU.

\subsection{ReLU$^2$: Best Function for LLM Deploying}

We present a summary of LLMs utilizing various activation functions by integrating the results of performance, sparsity, predictability, and hardware affinity above.
We deem an activation function as having comparable capabilities if the metric differential from the top-performing result is less than $1\%$.
For example, SwiGLU yields the best performance and ReGLU exhibits only a negligible performance difference, measured at $0.2\%$.
Therefore, we categorize ReGLU as having the attribute of high performance.

\begin{table}[t]
\centering
\small
\caption{Comparison of models with different activation functions.}
\label{tab:comparison}
\setlength{\tabcolsep}{1.5mm}{
\begin{tabular}{lcccc}
    \toprule  
    & SwiGLU& ReGLU & ReLU & ReLU$^2$\\
    \midrule  
    High Performance & $\checkmark$& $\checkmark$ &  & $\checkmark$ \\
    High Sparsity & & & $\checkmark$ & $\checkmark$ \\
    High Predictivity & & & $\checkmark$ & $\checkmark$ \\
    Hardware-friendly & & & $\checkmark$ & $\checkmark$ \\
    \bottomrule 
\end{tabular}
}
\vspace{-0.1in}
\end{table}

Table~\ref{tab:comparison} provides a comparative result of four 1B models with different activation functions.
Based on the result from the table, we observe that the ReLU$^2$ activation function achieves high results across all four evaluation aspects.
Based on the above results, we estimate that the 1B model with ReLU$^2$ can reduce $56\%$ of total computational cost in FLOPs based on the prediction sparsity with less than $1\%$ performance loss.
Besides, the I/O overhead of FFNs with ReLU$^2$ can be reduced by $92\%$ by utilizing both the sparsity ratio and the reuse ratio.

\section{Conclusion}

In this study, we propose a more general activation definition based on neuron output magnitudes with a magnitude threshold.
With this definition, we find that non-ReLU LLMs also exhibit sparse activation and observe sparser activation in ReLU-based LLMs than with the traditional definition.
All these sparse LLMs can be efficiently deployed with current frameworks of sparse computation~\cite{dejavu,powerinfer,flash}.
To find the most efficient activation function for sparse LLMs, we conduct thorough experiments on LLMs utilizing different activation functions, including ReLU, SwiGLU, ReGLU, and ReLU$^2$ and examine the sparsity of LLMs from three aspects: the trade-off between sparsity and performance, the predictivity of sparsity, and the hardware affinity.
The results indicate that models employing ReLU$^2$ excel across all three evaluation aspects, highlighting its potential as an efficient activation function for sparse LLMs.
We hope our work can provide a new perspective on the sparse computation of LLMs and facilitate future research to build more efficient LLMs.







\bibliography{example_paper}

\begin{thebibliography}{76}
\providecommand{\natexlab}[1]{#1}
\providecommand{\url}[1]{\texttt{#1}}
\expandafter\ifx\csname urlstyle\endcsname\relax
  \providecommand{\doi}[1]{doi: #1}\else
  \providecommand{\doi}{doi: \begingroup \urlstyle{rm}\Url}\fi

\bibitem[Ainslie et~al.(2023)Ainslie, Lee{-}Thorp, de~Jong, Zemlyanskiy, Lebr{\'{o}}n, and Sanghai]{GQA}
Ainslie, J., Lee{-}Thorp, J., de~Jong, M., Zemlyanskiy, Y., Lebr{\'{o}}n, F., and Sanghai, S.
\newblock {GQA:} training generalized multi-query transformer models from multi-head checkpoints.
\newblock In \emph{Proceedings of EMNLP}, pp.\  4895--4901, 2023.

\bibitem[Alizadeh et~al.(2023)Alizadeh, Mirzadeh, Belenko, Khatamifard, Cho, Mundo, Rastegari, and Farajtabar]{flash}
Alizadeh, K., Mirzadeh, I., Belenko, D., Khatamifard, K., Cho, M., Mundo, C. C.~D., Rastegari, M., and Farajtabar, M.
\newblock {LLM} in a flash: Efficient large language model inference with limited memory.
\newblock \emph{arXiv preprint arXiv:2312.11514}, 2023.

\bibitem[Almazrouei et~al.(2023)Almazrouei, Alobeidli, Alshamsi, Cappelli, Cojocaru, Debbah, Goffinet, Hesslow, Launay, Malartic, Mazzotta, Noune, Pannier, and Penedo]{Falcon}
Almazrouei, E., Alobeidli, H., Alshamsi, A., Cappelli, A., Cojocaru, R., Debbah, M., Goffinet, {\'{E}}., Hesslow, D., Launay, J., Malartic, Q., Mazzotta, D., Noune, B., Pannier, B., and Penedo, G.
\newblock The falcon series of open language models.
\newblock \emph{arxiv preprint arXiv:2311.16867}, 2023.

\bibitem[Aminabadi et~al.(2022)Aminabadi, Rajbhandari, Awan, Li, Li, Zheng, Ruwase, Smith, Zhang, Rasley, and He]{DBLP:conf/sc/AminabadiRALLZRSZRH22}
Aminabadi, R.~Y., Rajbhandari, S., Awan, A.~A., Li, C., Li, D., Zheng, E., Ruwase, O., Smith, S., Zhang, M., Rasley, J., and He, Y.
\newblock Deepspeed- inference: Enabling efficient inference of transformer models at unprecedented scale.
\newblock In \emph{Proceedings of SC}, 2022.

\bibitem[Artetxe et~al.(2022)Artetxe, Bhosale, Goyal, Mihaylov, Ott, Shleifer, Lin, Du, Iyer, Pasunuru, Anantharaman, Li, Chen, Akin, Baines, Martin, Zhou, Koura, O{'}Horo, Wang, Zettlemoyer, Diab, Kozareva, and Stoyanov]{artetxe-etal-2022-efficient}
Artetxe, M., Bhosale, S., Goyal, N., Mihaylov, T., Ott, M., Shleifer, S., Lin, X.~V., Du, J., Iyer, S., Pasunuru, R., Anantharaman, G., Li, X., Chen, S., Akin, H., Baines, M., Martin, L., Zhou, X., Koura, P.~S., O{'}Horo, B., Wang, J., Zettlemoyer, L., Diab, M., Kozareva, Z., and Stoyanov, V.
\newblock Efficient large scale language modeling with mixtures of experts.
\newblock In \emph{Proceedings of EMNLP}, pp.\  11699--11732, 2022.

\bibitem[Bengio(2013)]{DBLP:conf/slsp/Bengio13}
Bengio, Y.
\newblock Deep learning of representations: Looking forward.
\newblock In \emph{Proceedings of SLSP}, volume 7978, pp.\  1--37. Springer, 2013.

\bibitem[Bisk et~al.(2020)Bisk, Zellers, Bras, Gao, and Choi]{piqa}
Bisk, Y., Zellers, R., Bras, R.~L., Gao, J., and Choi, Y.
\newblock {PIQA:} reasoning about physical commonsense in natural language.
\newblock In \emph{Proceedings of AAAI}, 2020.

\bibitem[Bommasani et~al.(2021)Bommasani, Hudson, Adeli, Altman, Arora, von Arx, Bernstein, Bohg, Bosselut, Brunskill, et~al.]{bommasani2021opportunities}
Bommasani, R., Hudson, D.~A., Adeli, E., Altman, R., Arora, S., von Arx, S., Bernstein, M.~S., Bohg, J., Bosselut, A., Brunskill, E., et~al.
\newblock On the opportunities and risks of foundation models.
\newblock \emph{arXiv preprint arXiv:2108.07258}, 2021.

\bibitem[Breakspear(2017)]{breakspear2017dynamic}
Breakspear, M.
\newblock Dynamic models of large-scale brain activity.
\newblock \emph{Nature neuroscience}, 20\penalty0 (3):\penalty0 340--352, 2017.

\bibitem[Brown et~al.(2021)Brown, Mann, Ryder, Subbiah, Kaplan, Dhariwal, Neelakantan, Shyam, Sastry, Askell, Agarwal, Herbert-Voss, Krueger, Henighan, Child, Ramesh, Ziegler, Wu, Winter, Hesse, Chen, Sigler, Litwin, Gray, Chess, Clark, Berner, McCandlish, Radford, Sutskever, and Amodei]{GPT-3}
Brown, T.~B., Mann, B., Ryder, N., Subbiah, M., Kaplan, J., Dhariwal, P., Neelakantan, A., Shyam, P., Sastry, G., Askell, A., Agarwal, S., Herbert-Voss, A., Krueger, G., Henighan, T., Child, R., Ramesh, A., Ziegler, D.~M., Wu, J., Winter, C., Hesse, C., Chen, M., Sigler, E., Litwin, M., Gray, S., Chess, B., Clark, J., Berner, C., McCandlish, S., Radford, A., Sutskever, I., and Amodei, D.
\newblock Language models are {Few-Shot} learners.
\newblock In \emph{Proceedings of {NeurIPS}}, pp.\  1877--1901, 2021.

\bibitem[Bubeck et~al.(2023)Bubeck, Chandrasekaran, Eldan, Gehrke, Horvitz, Kamar, Lee, Lee, Li, Lundberg, Nori, Palangi, Ribeiro, and Zhang]{AGI-Sparks}
Bubeck, S., Chandrasekaran, V., Eldan, R., Gehrke, J., Horvitz, E., Kamar, E., Lee, P., Lee, Y.~T., Li, Y., Lundberg, S.~M., Nori, H., Palangi, H., Ribeiro, M.~T., and Zhang, Y.
\newblock Sparks of artificial general intelligence: Early experiments with {GPT-4}.
\newblock \emph{arXiv preprint arXiv:2303.12712}, 2023.

\bibitem[Cai et~al.(2024)Cai, Li, Geng, Peng, Lee, Chen, and Dao]{cai2024medusa}
Cai, T., Li, Y., Geng, Z., Peng, H., Lee, J.~D., Chen, D., and Dao, T.
\newblock Medusa: Simple llm inference acceleration framework with multiple decoding heads.
\newblock \emph{arXiv preprint arXiv: 2401.10774}, 2024.

\bibitem[Chen et~al.(2020)Chen, Medini, Farwell, Gobriel, Tai, and Shrivastava]{DBLP:conf/mlsys/ChenMFGTS20}
Chen, B., Medini, T., Farwell, J., Gobriel, S., Tai, T.~C., and Shrivastava, A.
\newblock {SLIDE} : In defense of smart algorithms over hardware acceleration for large-scale deep learning systems.
\newblock In \emph{Proceedings of MLSys}, 2020.

\bibitem[Chen et~al.(2021)Chen, Dao, Winsor, Song, Rudra, and R{\'{e}}]{DBLP:conf/nips/ChenDWSRR21}
Chen, B., Dao, T., Winsor, E., Song, Z., Rudra, A., and R{\'{e}}, C.
\newblock Scatterbrain: Unifying sparse and low-rank attention.
\newblock In \emph{Proceedings of NeurIPS}, 2021.

\bibitem[Chen et~al.(2023)Chen, Borgeaud, Irving, Lespiau, Sifre, and Jumper]{DBLP:journals/corr/abs-2302-01318}
Chen, C., Borgeaud, S., Irving, G., Lespiau, J., Sifre, L., and Jumper, J.
\newblock Accelerating large language model decoding with speculative sampling.
\newblock \emph{arXiv preprint arXiv:2302.01318}, 2023.

\bibitem[Cheng et~al.(2023)Cheng, Zhang, Shen, Cai, He, and Lv]{DBLP:journals/corr/abs-2309-05516}
Cheng, W., Zhang, W., Shen, H., Cai, Y., He, X., and Lv, K.
\newblock Optimize weight rounding via signed gradient descent for the quantization of llms.
\newblock \emph{arXiv preprint arXiv:2309.05516}, 2023.

\bibitem[Clark et~al.(2018)Clark, Cowhey, Etzioni, Khot, Sabharwal, Schoenick, and Tafjord]{arc}
Clark, P., Cowhey, I., Etzioni, O., Khot, T., Sabharwal, A., Schoenick, C., and Tafjord, O.
\newblock Think you have solved question answering? try arc, the {AI2} reasoning challenge.
\newblock \emph{arXiv preprint arXiv:1803.05457}, 2018.

\bibitem[Cobbe et~al.(2021)Cobbe, Kosaraju, Bavarian, Chen, Jun, Kaiser, Plappert, Tworek, Hilton, Nakano, Hesse, and Schulman]{GSM8k}
Cobbe, K., Kosaraju, V., Bavarian, M., Chen, M., Jun, H., Kaiser, L., Plappert, M., Tworek, J., Hilton, J., Nakano, R., Hesse, C., and Schulman, J.
\newblock Training verifiers to solve math word problems.
\newblock \emph{arXiv preprint arXiv:2110.14168}, 2021.

\bibitem[Dao et~al.(2022)Dao, Fu, Ermon, Rudra, and R{\'{e}}]{DBLP:conf/nips/DaoFERR22}
Dao, T., Fu, D.~Y., Ermon, S., Rudra, A., and R{\'{e}}, C.
\newblock Flashattention: Fast and memory-efficient exact attention with io-awareness.
\newblock In \emph{Proceedings of NeurIPS}, 2022.

\bibitem[Du et~al.(2022)Du, Huang, Dai, Tong, Lepikhin, Xu, Krikun, Zhou, Yu, Firat, Zoph, Fedus, Bosma, Zhou, Wang, Wang, Webster, Pellat, Robinson, Meier{-}Hellstern, Duke, Dixon, Zhang, Le, Wu, Chen, and Cui]{DBLP:conf/icml/DuHDTLXKZYFZFBZ22}
Du, N., Huang, Y., Dai, A.~M., Tong, S., Lepikhin, D., Xu, Y., Krikun, M., Zhou, Y., Yu, A.~W., Firat, O., Zoph, B., Fedus, L., Bosma, M.~P., Zhou, Z., Wang, T., Wang, Y.~E., Webster, K., Pellat, M., Robinson, K., Meier{-}Hellstern, K.~S., Duke, T., Dixon, L., Zhang, K., Le, Q.~V., Wu, Y., Chen, Z., and Cui, C.
\newblock Glam: Efficient scaling of language models with mixture-of-experts.
\newblock In \emph{Proceedings of ICML}, pp.\  5547--5569, 2022.

\bibitem[Elfwing et~al.(2018)Elfwing, Uchibe, and Doya]{SiLU}
Elfwing, S., Uchibe, E., and Doya, K.
\newblock Sigmoid-weighted linear units for neural network function approximation in reinforcement learning.
\newblock \emph{Neural Networks}, 107:\penalty0 3--11, 2018.

\bibitem[Fang et~al.(2021)Fang, Yu, Zhao, and Zhou]{DBLP:conf/ppopp/FangYZZ21}
Fang, J., Yu, Y., Zhao, C., and Zhou, J.
\newblock Turbotransformers: an efficient {GPU} serving system for transformer models.
\newblock In \emph{Proceedings of PPoPP}, 2021.

\bibitem[Fedus et~al.(2022)Fedus, Zoph, and Shazeer]{Switch-Transformer}
Fedus, W., Zoph, B., and Shazeer, N.
\newblock Switch transformers: Scaling to trillion parameter models with simple and efficient sparsity.
\newblock \emph{Journal of Machine Learning Research}, 23\penalty0 (120):\penalty0 1--39, 2022.

\bibitem[Frantar et~al.(2023)Frantar, Ashkboos, Hoefler, and Alistarh]{GPTQ}
Frantar, E., Ashkboos, S., Hoefler, T., and Alistarh, D.
\newblock {GPTQ}: Accurate quantization for generative pre-trained transformers.
\newblock In \emph{The Eleventh International Conference on Learning Representations}, 2023.

\bibitem[Gao et~al.(2021)Gao, Tow, Biderman, Black, DiPofi, Foster, Golding, Hsu, McDonell, Muennighoff, Phang, Reynolds, Tang, Thite, Wang, Wang, and Zou]{eval-harness}
Gao, L., Tow, J., Biderman, S., Black, S., DiPofi, A., Foster, C., Golding, L., Hsu, J., McDonell, K., Muennighoff, N., Phang, J., Reynolds, L., Tang, E., Thite, A., Wang, B., Wang, K., and Zou, A.
\newblock A framework for few-shot language model evaluation, September 2021.

\bibitem[Gu \& Dao(2023)Gu and Dao]{DBLP:journals/corr/abs-2312-00752}
Gu, A. and Dao, T.
\newblock Mamba: Linear-time sequence modeling with selective state spaces.
\newblock \emph{arXiv preprint arXiv:2312.00752}, 2023.

\bibitem[Han et~al.(2022)Han, Zeng, Zhao, Liu, Zhang, Zhou, Zhang, Chao, and Sun]{BMinf}
Han, X., Zeng, G., Zhao, W., Liu, Z., Zhang, Z., Zhou, J., Zhang, J., Chao, J., and Sun, M.
\newblock Bminf: An efficient toolkit for big model inference and tuning.
\newblock In \emph{Proceedings of ACL Demo}, pp.\  224--230, 2022.

\bibitem[Hendrycks \& Gimpel(2016)Hendrycks and Gimpel]{GeLU}
Hendrycks, D. and Gimpel, K.
\newblock Gaussian error linear units ({GELUs}).
\newblock \emph{arXiv preprint 1606.08415}, 2016.

\bibitem[Hendrycks et~al.(2021)Hendrycks, Burns, Basart, Zou, Mazeika, Song, and Steinhardt]{MMLU}
Hendrycks, D., Burns, C., Basart, S., Zou, A., Mazeika, M., Song, D., and Steinhardt, J.
\newblock Measuring massive multitask language understanding.
\newblock In \emph{Proceedings of ICLR}, 2021.

\bibitem[Kitaev et~al.(2020)Kitaev, Kaiser, and Levskaya]{DBLP:conf/iclr/KitaevKL20}
Kitaev, N., Kaiser, L., and Levskaya, A.
\newblock Reformer: The efficient transformer.
\newblock In \emph{Proceedings of ICLR}, 2020.

\bibitem[Kwon et~al.(2023)Kwon, Li, Zhuang, Sheng, Zheng, Yu, Gonzalez, Zhang, and Stoica]{vLLM}
Kwon, W., Li, Z., Zhuang, S., Sheng, Y., Zheng, L., Yu, C.~H., Gonzalez, J., Zhang, H., and Stoica, I.
\newblock Efficient memory management for large language model serving with pagedattention.
\newblock In \emph{Proceedings of SOSP}, pp.\  611--626, 2023.

\bibitem[Lepikhin et~al.(2021)Lepikhin, Lee, Xu, Chen, Firat, Huang, Krikun, Shazeer, and Chen]{GShard}
Lepikhin, D., Lee, H., Xu, Y., Chen, D., Firat, O., Huang, Y., Krikun, M., Shazeer, N., and Chen, Z.
\newblock Gshard: Scaling giant models with conditional computation and automatic sharding.
\newblock In \emph{Proceedings of ICLR}, 2021.

\bibitem[Leviathan et~al.(2023)Leviathan, Kalman, and Matias]{DBLP:conf/icml/LeviathanKM23}
Leviathan, Y., Kalman, M., and Matias, Y.
\newblock Fast inference from transformers via speculative decoding.
\newblock In \emph{Proceedings of ICML}, 2023.

\bibitem[Li et~al.(2023)Li, You, Bhojanapalli, Li, Rawat, Reddi, Ye, Chern, Yu, Guo, and Kumar]{li2023the}
Li, Z., You, C., Bhojanapalli, S., Li, D., Rawat, A.~S., Reddi, S.~J., Ye, K., Chern, F., Yu, F., Guo, R., and Kumar, S.
\newblock The lazy neuron phenomenon: On emergence of activation sparsity in transformers.
\newblock In \emph{Proceedings of ICLR}, 2023.

\bibitem[Lin et~al.(2023)Lin, Tang, Tang, Yang, Dang, and Han]{AWQ}
Lin, J., Tang, J., Tang, H., Yang, S., Dang, X., and Han, S.
\newblock {AWQ:} activation-aware weight quantization for {LLM} compression and acceleration.
\newblock \emph{arXiv preprint arXiv:2306.00978}, 2023.

\bibitem[Lin et~al.(2022)Lin, Hilton, and Evans]{truthfulqa}
Lin, S., Hilton, J., and Evans, O.
\newblock Truthfulqa: Measuring how models mimic human falsehoods.
\newblock In \emph{Proceedings of ACL}, 2022.

\bibitem[Liu et~al.(2023)Liu, Wang, Dao, Zhou, Yuan, Song, Shrivastava, Zhang, Tian, R{\'{e}}, and Chen]{dejavu}
Liu, Z., Wang, J., Dao, T., Zhou, T., Yuan, B., Song, Z., Shrivastava, A., Zhang, C., Tian, Y., R{\'{e}}, C., and Chen, B.
\newblock Deja vu: Contextual sparsity for efficient llms at inference time.
\newblock In Krause, A., Brunskill, E., Cho, K., Engelhardt, B., Sabato, S., and Scarlett, J. (eds.), \emph{Preceedings of ICML}, volume 202, pp.\  22137--22176, 2023.

\bibitem[Loshchilov \& Hutter(2017)Loshchilov and Hutter]{loshchilov2017decoupled}
Loshchilov, I. and Hutter, F.
\newblock Decoupled weight decay regularization.
\newblock \emph{arXiv preprint arXiv:1711.05101}, 2017.

\bibitem[Michel et~al.(2019)Michel, Levy, and Neubig]{DBLP:conf/nips/MichelLN19}
Michel, P., Levy, O., and Neubig, G.
\newblock Are sixteen heads really better than one?
\newblock In \emph{Proceedings of NeurIPS}, 2019.

\bibitem[Mihaylov et~al.(2018)Mihaylov, Clark, Khot, and Sabharwal]{openbookqa}
Mihaylov, T., Clark, P., Khot, T., and Sabharwal, A.
\newblock Can a suit of armor conduct electricity? {A} new dataset for open book question answering.
\newblock In \emph{Proceedings of EMNLP}, 2018.

\bibitem[Mirzadeh et~al.(2023)Mirzadeh, Alizadeh, Mehta, Mundo, Tuzel, Samei, Rastegari, and Farajtabar]{mirzadeh2023relu}
Mirzadeh, I., Alizadeh, K., Mehta, S., Mundo, C. C.~D., Tuzel, O., Samei, G., Rastegari, M., and Farajtabar, M.
\newblock Relu strikes back: Exploiting activation sparsity in large language models.
\newblock \emph{arXiv preprint arXiv:2310.04564}, 2023.

\bibitem[Mosaicml(2023)]{llmfoundry}
Mosaicml.
\newblock { Llm foundry}.
\newblock \url{https://github.com/mosaicml/llm-foundry}, 2023.

\bibitem[Nair \& Hinton(2010)Nair and Hinton]{ReLU}
Nair, V. and Hinton, G.~E.
\newblock Rectified linear units improve restricted boltzmann machines.
\newblock In \emph{Proceedings of ICML}, pp.\  807--814, 2010.

\bibitem[OpenAI(2023)]{GPT-4}
OpenAI.
\newblock {GPT-4} technical report.
\newblock \emph{arxiv preprint arXiv:2303.08774}, 2023.

\bibitem[Ouyang et~al.(2022)Ouyang, Wu, Jiang, Almeida, Wainwright, Mishkin, Zhang, Agarwal, Slama, Ray, Schulman, Hilton, Kelton, Miller, Simens, Askell, Welinder, Christiano, Leike, and Lowe]{InstructGPT}
Ouyang, L., Wu, J., Jiang, X., Almeida, D., Wainwright, C.~L., Mishkin, P., Zhang, C., Agarwal, S., Slama, K., Ray, A., Schulman, J., Hilton, J., Kelton, F., Miller, L., Simens, M., Askell, A., Welinder, P., Christiano, P.~F., Leike, J., and Lowe, R.
\newblock Training language models to follow instructions with human feedback.
\newblock In \emph{Proceedings of NeurIPS}, 2022.

\bibitem[Paperno et~al.(2016)Paperno, Kruszewski, Lazaridou, Pham, Bernardi, Pezzelle, Baroni, Boleda, and Fern{\'{a}}ndez]{lambada}
Paperno, D., Kruszewski, G., Lazaridou, A., Pham, Q.~N., Bernardi, R., Pezzelle, S., Baroni, M., Boleda, G., and Fern{\'{a}}ndez, R.
\newblock The {LAMBADA} dataset: Word prediction requiring a broad discourse context.
\newblock In \emph{Proceedings of ACL}, 2016.

\bibitem[Pariz et~al.(2021)Pariz, Fischer, Valizadeh, and Mirasso]{pariz2021transmission}
Pariz, A., Fischer, I., Valizadeh, A., and Mirasso, C.
\newblock Transmission delays and frequency detuning can regulate information flow between brain regions.
\newblock \emph{PLoS computational biology}, 17\penalty0 (4):\penalty0 e1008129, 2021.

\bibitem[Penedo et~al.(2023)Penedo, Malartic, Hesslow, Cojocaru, Cappelli, Alobeidli, Pannier, Almazrouei, and Launay]{refinedweb}
Penedo, G., Malartic, Q., Hesslow, D., Cojocaru, R., Cappelli, A., Alobeidli, H., Pannier, B., Almazrouei, E., and Launay, J.
\newblock The {R}efined{W}eb dataset for {F}alcon {LLM}: outperforming curated corpora with web data, and web data only.
\newblock \emph{arXiv preprint arXiv:2306.01116}, 2023.

\bibitem[Peng et~al.(2023)Peng, Alcaide, Anthony, Albalak, Arcadinho, Biderman, Cao, Cheng, Chung, Derczynski, Du, Grella, Gv, He, Hou, Kazienko, Kocon, Kong, Koptyra, Lau, Lin, Mantri, Mom, Saito, Song, Tang, Wind, Wozniak, Zhang, Zhou, Zhu, and Zhu]{DBLP:conf/emnlp/PengAAAABCCCDDG23}
Peng, B., Alcaide, E., Anthony, Q., Albalak, A., Arcadinho, S., Biderman, S., Cao, H., Cheng, X., Chung, M., Derczynski, L., Du, X., Grella, M., Gv, K., He, X., Hou, H., Kazienko, P., Kocon, J., Kong, J., Koptyra, B., Lau, H., Lin, J., Mantri, K. S.~I., Mom, F., Saito, A., Song, G., Tang, X., Wind, J.~S., Wozniak, S., Zhang, Z., Zhou, Q., Zhu, J., and Zhu, R.
\newblock {RWKV:} reinventing rnns for the transformer era.
\newblock In \emph{Findings of EMNLP}, 2023.

\bibitem[Pope et~al.(2022)Pope, Douglas, Chowdhery, Devlin, Bradbury, Levskaya, Heek, Xiao, Agrawal, and Dean]{DBLP:journals/corr/abs-2211-05102}
Pope, R., Douglas, S., Chowdhery, A., Devlin, J., Bradbury, J., Levskaya, A., Heek, J., Xiao, K., Agrawal, S., and Dean, J.
\newblock Efficiently scaling transformer inference.
\newblock \emph{arXiv preprint arXiv:2211.05102}, 2022.

\bibitem[Raffel et~al.(2020)Raffel, Shazeer, Roberts, Lee, Narang, Matena, Zhou, Li, and Liu]{T5}
Raffel, C., Shazeer, N., Roberts, A., Lee, K., Narang, S., Matena, M., Zhou, Y., Li, W., and Liu, P.~J.
\newblock Exploring the limits of transfer learning with a unified {Text-to-Text} transformer.
\newblock \emph{J. Mach. Learn. Res.}, 21:\penalty0 140:1--140:67, 2020.

\bibitem[Rajbhandari et~al.(2022)Rajbhandari, Li, Yao, Zhang, Aminabadi, Awan, Rasley, and He]{DBLP:conf/icml/RajbhandariLYZA22}
Rajbhandari, S., Li, C., Yao, Z., Zhang, M., Aminabadi, R.~Y., Awan, A.~A., Rasley, J., and He, Y.
\newblock Deepspeed-moe: Advancing mixture-of-experts inference and training to power next-generation {AI} scale.
\newblock In \emph{Proceedings of ICML}, 2022.

\bibitem[Riquelme et~al.(2021)Riquelme, Puigcerver, Mustafa, Neumann, Jenatton, Pinto, Keysers, and Houlsby]{DBLP:conf/nips/RiquelmePMNJPKH21}
Riquelme, C., Puigcerver, J., Mustafa, B., Neumann, M., Jenatton, R., Pinto, A.~S., Keysers, D., and Houlsby, N.
\newblock Scaling vision with sparse mixture of experts.
\newblock In \emph{Proceedings of NeurIPS}, 2021.

\bibitem[Sakaguchi et~al.(2020)Sakaguchi, Bras, Bhagavatula, and Choi]{winogrande}
Sakaguchi, K., Bras, R.~L., Bhagavatula, C., and Choi, Y.
\newblock Winogrande: An adversarial winograd schema challenge at scale.
\newblock In \emph{Proceedings of AAAI}, pp.\  8732--8740, 2020.

\bibitem[Shazeer(2020)]{GLU}
Shazeer, N.
\newblock {GLU} variants improve transformer.
\newblock \emph{arXiv preprint arXiv:2002.05202}, 2020.

\bibitem[Sheng et~al.(2023)Sheng, Zheng, Yuan, Li, Ryabinin, Chen, Liang, R{\'{e}}, Stoica, and Zhang]{Flexgen}
Sheng, Y., Zheng, L., Yuan, B., Li, Z., Ryabinin, M., Chen, B., Liang, P., R{\'{e}}, C., Stoica, I., and Zhang, C.
\newblock Flexgen: High-throughput generative inference of large language models with a single {GPU}.
\newblock In \emph{Proceedings of ICML}, pp.\  31094--31116, 2023.

\bibitem[So et~al.(2021)So, Manke, Liu, Dai, Shazeer, and Le]{relu2}
So, D.~R., Manke, W., Liu, H., Dai, Z., Shazeer, N., and Le, Q.~V.
\newblock Primer: Searching for efficient transformers for language modeling.
\newblock \emph{arXiv preprint arXiv:2109.08668}, 2021.

\bibitem[Soboleva et~al.(2023)Soboleva, Al-Khateeb, Myers, Steeves, Hestness, and Dey]{cerebras2023slimpajama}
Soboleva, D., Al-Khateeb, F., Myers, R., Steeves, J.~R., Hestness, J., and Dey, N.
\newblock {SlimPajama: A 627B token cleaned and deduplicated version of RedPajama}, 2023.

\bibitem[Song et~al.(2023)Song, Mi, Xie, and Chen]{powerinfer}
Song, Y., Mi, Z., Xie, H., and Chen, H.
\newblock Powerinfer: Fast large language model serving with a consumer-grade gpu.
\newblock \emph{arXiv preprint arXiv:2312.12456}, 2023.

\bibitem[Spector \& R{\'{e}}(2023)Spector and R{\'{e}}]{DBLP:journals/corr/abs-2308-04623}
Spector, B. and R{\'{e}}, C.
\newblock Accelerating {LLM} inference with staged speculative decoding.
\newblock \emph{arXiv preprint arXiv:2308.04623}, 2023.

\bibitem[Su et~al.(2024)Su, Ahmed, Lu, Pan, Bo, and Liu]{su2024roformer}
Su, J., Ahmed, M., Lu, Y., Pan, S., Bo, W., and Liu, Y.
\newblock Roformer: Enhanced transformer with rotary position embedding.
\newblock \emph{Neurocomputing}, 568:\penalty0 127063, 2024.

\bibitem[Touvron et~al.(2023{\natexlab{a}})Touvron, Lavril, Izacard, Martinet, Lachaux, Lacroix, Rozi{\`{e}}re, Goyal, Hambro, Azhar, Rodriguez, Joulin, Grave, and Lample]{llama1}
Touvron, H., Lavril, T., Izacard, G., Martinet, X., Lachaux, M., Lacroix, T., Rozi{\`{e}}re, B., Goyal, N., Hambro, E., Azhar, F., Rodriguez, A., Joulin, A., Grave, E., and Lample, G.
\newblock Llama: Open and efficient foundation language models.
\newblock \emph{arxiv preprint arXiv:2302.13971}, 2023{\natexlab{a}}.

\bibitem[Touvron et~al.(2023{\natexlab{b}})Touvron, Martin, Stone, Albert, et~al.]{LLAMA2}
Touvron, H., Martin, L., Stone, K., Albert, P., et~al.
\newblock Llama 2: Open foundation and fine-tuned chat models.
\newblock \emph{arxiv preprint arXiv:2307.09288}, 2023{\natexlab{b}}.

\bibitem[van~den Brand et~al.(2023)van~den Brand, Song, and Zhou]{DBLP:journals/corr/abs-2304-02207}
van~den Brand, J., Song, Z., and Zhou, T.
\newblock Algorithm and hardness for dynamic attention maintenance in large language models.
\newblock \emph{arXiv preprint arXiv:2304.02207}, 2023.

\bibitem[Vaswani et~al.(2017)Vaswani, Shazeer, Parmar, and Uszkoreit]{Transformer}
Vaswani, A., Shazeer, N., Parmar, N., and Uszkoreit, J.
\newblock Attention is all you need.
\newblock In \emph{Proceedings of {NeurIPS}}, pp.\  5998--6008, 2017.

\bibitem[Wang et~al.(2021)Wang, Xiong, Wei, Wang, and Li]{DBLP:conf/naacl/WangXWWL21}
Wang, X., Xiong, Y., Wei, Y., Wang, M., and Li, L.
\newblock Lightseq: {A} high performance inference library for transformers.
\newblock In \emph{Proceedings of NAACL-HLT}, 2021.

\bibitem[Wang et~al.(2022)Wang, Wen, Zhang, Hou, Liu, and Li]{SkillNeuron}
Wang, X., Wen, K., Zhang, Z., Hou, L., Liu, Z., and Li, J.
\newblock Finding skill neurons in pre-trained transformer-based language models.
\newblock In \emph{Proceedings of EMNLP}, 2022.

\bibitem[Xiao et~al.(2023)Xiao, Lin, Seznec, Wu, Demouth, and Han]{DBLP:conf/icml/XiaoLSWDH23}
Xiao, G., Lin, J., Seznec, M., Wu, H., Demouth, J., and Han, S.
\newblock Smoothquant: Accurate and efficient post-training quantization for large language models.
\newblock In \emph{Proceedings of ICML}, 2023.

\bibitem[Yao et~al.(2022)Yao, Aminabadi, Zhang, Wu, Li, and He]{DBLP:conf/nips/YaoAZWLH22}
Yao, Z., Aminabadi, R.~Y., Zhang, M., Wu, X., Li, C., and He, Y.
\newblock Zeroquant: Efficient and affordable post-training quantization for large-scale transformers.
\newblock In \emph{Proceedings of NeurIPS}, 2022.

\bibitem[Yu et~al.(2022)Yu, Jeong, Kim, Kim, and Chun]{DBLP:conf/osdi/YuJKKC22}
Yu, G., Jeong, J.~S., Kim, G., Kim, S., and Chun, B.
\newblock Orca: {A} distributed serving system for transformer-based generative models.
\newblock In \emph{Proceedings of OSDI}, 2022.

\bibitem[Zellers et~al.(2019)Zellers, Holtzman, Bisk, Farhadi, and Choi]{hellaswag}
Zellers, R., Holtzman, A., Bisk, Y., Farhadi, A., and Choi, Y.
\newblock Hellaswag: Can a machine really finish your sentence?
\newblock In \emph{Proceedings of ACL}, 2019.

\bibitem[Zhang \& Sennrich(2019)Zhang and Sennrich]{RMSNorm}
Zhang, B. and Sennrich, R.
\newblock Root mean square layer normalization.
\newblock In \emph{Proceedings of NeurIPS}, 2019.

\bibitem[Zhang et~al.(2022{\natexlab{a}})Zhang, Roller, Goyal, Artetxe, Chen, Chen, Dewan, Diab, Li, Lin, Mihaylov, Ott, Shleifer, Shuster, Simig, Koura, Sridhar, Wang, and Zettlemoyer]{OPT}
Zhang, S., Roller, S., Goyal, N., Artetxe, M., Chen, M., Chen, S., Dewan, C., Diab, M.~T., Li, X., Lin, X.~V., Mihaylov, T., Ott, M., Shleifer, S., Shuster, K., Simig, D., Koura, P.~S., Sridhar, A., Wang, T., and Zettlemoyer, L.
\newblock {OPT:} open pre-trained transformer language models.
\newblock \emph{arXiv preprint arXiv:2205.01068}, 2022{\natexlab{a}}.

\bibitem[Zhang et~al.(2022{\natexlab{b}})Zhang, Lin, Liu, Li, Sun, and Zhou]{zhang2022moefication}
Zhang, Z., Lin, Y., Liu, Z., Li, P., Sun, M., and Zhou, J.
\newblock {MoEfication}: Transformer feed-forward layers are mixtures of experts.
\newblock In \emph{Findings of ACL}, 2022{\natexlab{b}}.

\bibitem[Zhang et~al.(2023)Zhang, Zeng, Lin, Xiao, Wang, Han, Liu, Xie, Sun, and Zhou]{zhang2023emergent}
Zhang, Z., Zeng, Z., Lin, Y., Xiao, C., Wang, X., Han, X., Liu, Z., Xie, R., Sun, M., and Zhou, J.
\newblock Emergent modularity in pre-trained transformers.
\newblock In \emph{Findings of ACL}, 2023.

\bibitem[Zhao et~al.(2023)Zhao, Zhou, Li, Tang, Wang, Hou, Min, Zhang, Zhang, Dong, et~al.]{zhao2023survey}
Zhao, W.~X., Zhou, K., Li, J., Tang, T., Wang, X., Hou, Y., Min, Y., Zhang, B., Zhang, J., Dong, Z., et~al.
\newblock A survey of large language models.
\newblock \emph{arXiv preprint arXiv:2303.18223}, 2023.

\end{thebibliography}
\bibliographystyle{icml2024}

\newpage
\appendix
\onecolumn

\section{Experimental Details}
\label{app:experimental_details}

\subsection{Evaluation Datasets}

For the evaluation of LLaMA-2 7B, we use MMLU~\cite{MMLU}, the challenging set of ARC~\cite{arc}, Winogrande~\cite{winogrande}, HellaSwag~\cite{hellaswag}, TruthfulQA~\cite{truthfulqa}, and GSM8k~\cite{GSM8k} from Open LLM Leaderboard\footnote{\url{https://huggingface.co/open-llm-leaderboard}} and follow the evaluation protocol set by the leaderboard using Language Model Evaluation Harness~\cite{eval-harness}.
For TruthfulQA, the lower the accuracy, the better the performance.
Hence, when calculating the average accuracy across all datasets, we use $1$ minus the accuracy of TruthfulQA as the performance on TruthfulQA.
We report the average accuracy across all datasets as the final performance.

For the evaluation of 1B models, we use the following datasets: both the easy and challenging set of ARC~\cite{arc}, Winogrande~\cite{winogrande}, HellaSwag~\cite{hellaswag}, TruthfulQA~\cite{truthfulqa}, LAMBADA~\cite{lambada}, PIQA~\cite{piqa}, and OpenBookQA~\cite{openbookqa}.
We exclude MMLU and GSM8k from the evaluation because the 1B models do not have enough capacity to achieve reasonable performance on these two datasets.
We conduct zero-shot evaluation on LAMBADA, PIQA, and OpenBookQA.
The rest of the datasets are evaluated using the same evaluation protocol as the evaluation of LLaMA-2 7B.

For the corpus used for calculating the sparsity statistics, we a mixture of texts from Wikipedia, Pile, and Common Crawl.
The total number of tokens is about $2^{17}$.
Since we need to store the activation values of all neurons, we only use a part of the corpus to calculate the sparsity statistics.

\subsection{Details of Predictor}

Specifically, the predictor is a two-layer neural network computed by
\begin{equation}
    \hat{\vy} = \sigma(\mW_2(\mW_1\vx+\vb_1)+\vb_2),
\end{equation}
where $\mW_1\in\mathbb{R}^{d_{\text{model}}\times r}$, $\vb_1\in\mathbb{R}^{r}$, $\mW_2\in\mathbb{R}^{d_{\text{ff}}\times r}$, $\vb_2\in\mathbb{R}^{d_{ff}}$, $\sigma$ is the sigmoid function, and $r$ is a hyperparameter which is much smaller than $d_{\text{model}}$ and $d_{\text{ff}}$ to ensure the efficiency of the predictor.
$\hat{\vy}$ is the activation prediction score of the FFN, and $\hat{\vy}_i$ is the prediction score of the $i$-th neuron.
If $\hat{\vy}_i$ is larger than $0.5$, the $i$-th neuron is predicted to be activated; otherwise, it is predicted to be inactive.
Based on the activation records of each FFN, we construct the training set and validation set for the predictor.
The dataset contains the FFN input $\vx$, the activation records $\vy$, where $\vy_i=1$ if the $i$-th neuron output magnitude is larger than the threshold, and $\vy_i=0$ otherwise.
The training objective is to minimize the binary cross-entropy loss between the prediction score $\hat{\vy}$ and the activation records $\vy$.

\subsection{Training of 1.3B Model}

In this subsection, we will introduce the details of training the 1.3B model, including the model architecture, types of data used, and hyperparameters. The evaluation results of the final 1.3B models are shown in Table~\ref{tab:1B_average}.

\begin{table}[h]
\centering
\small
\caption{Accuracy (\%) of 1B models on the evaluation datasets with average. ``ARC:E'' and ``ARC:C'' refer to the easy and challenging set of ARC, respectively.}
\label{tab:1B_average}
\begin{tabular}{lrrrrrrrrr}
\toprule
         & \multicolumn{1}{l}{ARC:E} & \multicolumn{1}{l}{ARC:C} & \multicolumn{1}{l}{HellaSwag} & \multicolumn{1}{l}{Winograde} & \multicolumn{1}{l}{TruthfulQA} & \multicolumn{1}{l}{LAMBADA} & \multicolumn{1}{l}{PIQA} & \multicolumn{1}{l}{OpenBookQA} & \multicolumn{1}{l}{Average}  \\ \midrule
SwiGLU   & 64.86                        & 34.04                             & 42.46                         & 51.93                         & 62.29                          & 50.57                       & 73.94                    & 24.20                          & 50.53                     \\
ReLU     & 61.62                        & 30.72                             & 40.79                         & 53.75                         & 62.68                          & 47.74                       & 72.91                    & 20.60                         & 48.85                     \\ 
ReGLU    & 63.01                        & 33.02                             & 41.72                         & 55.64                         & 63.23                          & 49.95                       & 75.14                    & 21.20                          & 50.36                     \\
ReLU$^2$ & 64.65                        & 31.66                             & 41.93                         & 54.46                         & 64.05                          & 49.62                       & 74.10                    & 23.40                          & 50.48                     \\\bottomrule
\end{tabular}
\end{table}

\subsubsection{Architecture}
We adopt a similar model architecture to LLaMA-2~\cite{LLAMA2} with the following details:

\begin{table}[htbp]
    \centering
    \normalsize
    \caption{ 
      Details of model architecture.
    }
    \scalebox{1.0}{
    \begin{tabular}{lcccccccc}
      \toprule
      \textbf{Hidden size}  & \textbf{Context Len}& \textbf{Heads} & \textbf{Layers} & \textbf{Vocab size}\\
  
      \midrule    
      2,048       & 2,048       & 32    & 24  & 32,000  \\
      \bottomrule
    \end{tabular}
    }
    \label{tab:model}
\end{table}

\paragraph{Activation Function and Intermediate Hidden Size} 
We focus on four different activation functions: ReLU~\cite{ReLU}, ReLU$^2$~\cite{relu2}, ReGLU~\cite{GLU}, SwiGLU~\cite{GLU}. 
We set different intermediate hidden sizes for different activation functions to ensure the same number of parameters.
For ReLU and ReLU$^2$, we set the intermediate hidden size to 8192. For ReGLU and SwiGLU, we set it to 5376.

\paragraph{Multi-Head Attention} 
For Attention block, we adopt the LLaMA2-7B's architecture, apply pre-normalization using RMSNorm~\cite{RMSNorm} and RoPE~\cite{su2024roformer} for Positional embedding.
\subsubsection{Pre-training Data}
We adopt a mixture of natural language data from SlimPajama~\cite{cerebras2023slimpajama} and refinedweb~\cite{refinedweb}.
After sampling and combining these two corpora, we have 100 billion tokens for pre-training in total. 
These 1B models are trained on these tokens for one epoch.
\subsubsection{Training Hyperparameters}
We utilize llm-foundry as our training framework~\cite{llmfoundry}.
Our models are trained using AdamW optimizer~\cite{loshchilov2017decoupled}, with the following hyper-parameters: $\beta_1 = 0.9, \beta_2 = 0.95$.
We adopt a cosine learning rate schedule and we use the default weight decay and gradient clipping hyper-parameters. (see Table ~\ref{tab:hyperparameters} for more details).
\begin{table}[htbp]
    \centering
    \normalsize
    \caption{ 
      Details of optimization hyper-parameters.
    }
    \scalebox{1.0}{
    \begin{tabular}{lcccccccc}
      \toprule
      \textbf{Sequence Length}  & \textbf{Batch Size}& \textbf{Learning Rate} & \textbf{Warmup Steps} & \textbf{Hardware}\\
  
      \midrule    
      2,048       & 2,048       & 3e$^{-4}$    & 2000  & 64 A100-80G GPUs  \\
      \bottomrule
    \end{tabular}
    }
    \label{tab:hyperparameters}
\end{table}

\subsection{Further Pre-training of LLaMA using ReGLU}

To study the scaling effect of model size, we further pre-train the LLaMA models using ReGLU with different sizes.
To accelerate the pre-training process, we use the parameters of the original LLaMA-2 models as the initialization of the ReGLU-based LLaMA models.
The activation function is the only difference between the original LLaMA-2 models and the ReGLU-based LLaMA models.
Specifically, we set the batch size to $2048$ for 7B and 13B models, and $1920$ for 70B models.
The learning rate is set to $3e^{-5}$ for all models.
The training data includes Wikipedia, Pile, and StackOverflow.
The total number of tokens is about 5 billion.
We use 8 A100-80G GPUs for pre-training the 7B, 16 A100-80G GPUs for pre-training the 13B, and 40 A100-80G GPUs for pre-training the 70B.

The performance of the pre-trained models on the evaluation datasets is shown in Table~\ref{tab:llama-pretrain}. From the table, we can see that the ReGLU-based LLaMA models achieve comparable performance with the original LLaMA-2 models under the sizes of 7B and 13B.
However, the performance of the ReGLU-based LLaMA model lags behind the original LLaMA-2 model under the size of 70B by about $6\%$ on average.
We conjecture that the performance gap is caused by the insufficient pre-training data.
We believe that the performance of the ReGLU-based LLaMA model can be further improved if we add more pre-training data.

\begin{table}[]
\centering
\caption{Performance (\%) of LLaMA with different sizes and activation functions on the evaluation datasets.}
\label{tab:llama-pretrain}
\begin{tabular}{llrrrrrrr}
\toprule
                     &        & \multicolumn{1}{l}{ARC} & \multicolumn{1}{l}{HellaSwag} & \multicolumn{1}{l}{MMLU} & \multicolumn{1}{l}{TruthfulQA} & \multicolumn{1}{l}{Winogrande} & \multicolumn{1}{l}{GSM8K} & \multicolumn{1}{l}{Average} \\ \midrule
\multirow{2}{*}{7B}  & SwiGLU    & 53.07                   & 78.59                         & 46.87                    & 61.24                          & 74.03                          & 14.48                     & 54.71 \\
& ReGLU   & 49.48                   & 74.67                         & 44.84                    & 60.96                          & 69.37                          & 10.61                        & 51.66                               \\ 
\midrule
\multirow{2}{*}{13B} & SwiGLU &   59.30                    & 82.15                         & 55.67                    & 62.61                          & 76.64                          & 10.84                     &   57.87    \\
& ReGLU  &  53.92                   & 78.22                         & 52.10                     & 62.64                          & 73.40                           & 16.37                     & 56.11                                    \\ \midrule
\multirow{2}{*}{70B} & SwiGLU & 67.32                   & 87.33                         & 69.83                    & 55.08                          & 83.74                          & 54.06                     & 69.56       \\
& ReGLU  & 62.79                   & 84.00                            & 63.39                    & 53.40                           & 80.97                          & 36.31                     & 63.48                            \\ \bottomrule          
\end{tabular}
\end{table}

\section{Neuron Decomposition of FFNs}
\label{app:neuron_decomposition}

Different from SwiGLU adopted by LLaMA, the original Transformer~\cite{Transformer} adopts a two-layer perceptron as FFN, which is also used in current LLMs such as Falcon~\cite{Falcon}.
Specifically, a two-layer FFN is computed by
\begin{equation}
    \text{FFN}(x) = \mW^{out} \sigmoid(\mW^{in} \vx + \vb^{in}) + \vb^{out},
\end{equation}
where $\mW^{in} \in \R^{d_{\text{ff}} \times d_{\text{model}}}$, $\vb^{in} \in \R^{d_{\text{ff}}}$, $\mW^{out} \in \R^{d_{\text{model}} \times d_{\text{ff}}}$, $\vb^{out} \in \R^{d_{\text{model}}}$, $\sigmoid$ is the activation function, $d_{\text{model}}$ is the dimension of the input and output representations, and $d_{\text{ff}}$ is the dimension of the hidden layer.
For simplicity, we omit the bias terms in the following discussion.
We decompose the FFN along the dimension of $d_{\text{ff}}$ and get $d_{\text{ff}}$ neurons, the $i$-th of which is computed by
\begin{equation}
    \text{n}_i(x) = \mW^{out}_{:,i} \sigmoid(\mW^{in}_{i,:} \vx),
\end{equation}
where $\mW^{in}_{i,:}$ and $\mW^{out}_{:,i}$ are the $i$-th row and column of $\mW^{in}$ and $\mW^{out}$, respectively.
We denote the activation value of the $i$-th neuron as $a_i(x) = \sigmoid(\mW^{in}_{i,:} \vx)$, which is a scalar, and the output representation of the $i$-th neuron as $\text{n}_i(x)$, which is a vector.
The output representation of the FFN is the summation of the output representations of all neurons, i.e., $\text{FFN}(x) = \sum_{i=1}^{d_{\text{ff}}} \text{n}_i(x)$.

\section{Threshold-finding Method}
\label{app:threshold_search}

\begin{algorithm}[tb]
    \caption{Adaptive Threshold Finding for Sparse Activation}
    \label{alg:adaptiveThreshold}
 \begin{algorithmic}
    \STATE {\bfseries Input:} neuron outputs $neuronOutputs$, CETT upper bound $CETTUpperBound$
    \STATE {\bfseries Output:} best threshold $bestThreshold$

    \STATE $candidateThresholds \gets$ CalculateQuantiles($neuronOutputs$)
    \STATE $low \gets 0$
    \STATE $high \gets$ length of $candidateThresholds - 1$
    \STATE $bestThreshold \gets 0$

    \WHILE{$low \leq high$}
        \STATE $mid \gets \lfloor (low + high) / 2 \rfloor$
        \STATE $threshold \gets candidateThresholds[mid]$
        \STATE $CETT \gets$ CalculateCETT($neuronOutputs, threshold$)

        \IF{$CETT \leq CETTUpperBound$}
            \STATE $bestThreshold \gets threshold$
            \STATE $low \gets mid + 1$
        \ELSE
            \STATE $high \gets mid - 1$
        \ENDIF
    \ENDWHILE

    \STATE \textbf{return} $bestThreshold$



 \end{algorithmic}

 \end{algorithm}

In this section, we introduce the threshold search method used in our experiments.
The threshold search method is used to find the best threshold based on the given neuron outputs and the CETT upper bound.
The algorithm is shown in \cref{alg:adaptiveThreshold}.
CalculateQuantiles is used to calculate the quantiles of the neuron output magnitudes and usually returns 1000 quantiles.
CalculateCETT is used to calculate the CETT based on the given threshold and the neuron outputs. In the experiments, the CETT upper bound is chosen from $\{0.01, 0.02, 0.04, 0.1, 0.2, 0.3, 0.4, 0.5\}$.

\section{Hot-activated Neurons}

\begin{figure}
\centering
\includegraphics[width=0.9\linewidth]{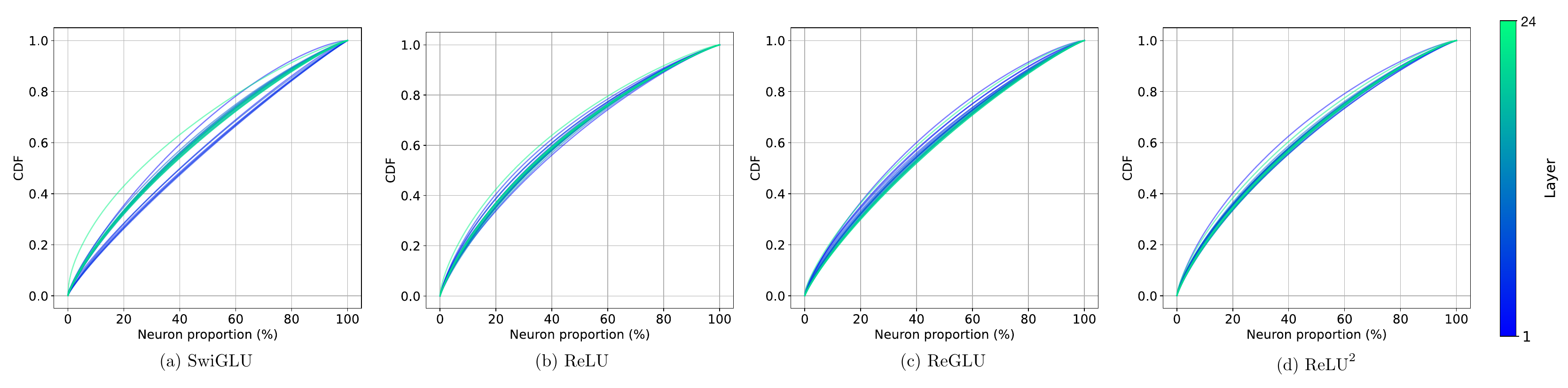}
\caption{Cumulative distribution function (CDF) of neuron activation for 1B models with different activation functions. The x-axis is the neuron proportion (sorted in descending order).}
\label{fig:commonality}
\end{figure}

\begin{figure}
\centering
\includegraphics[width=0.9\linewidth]{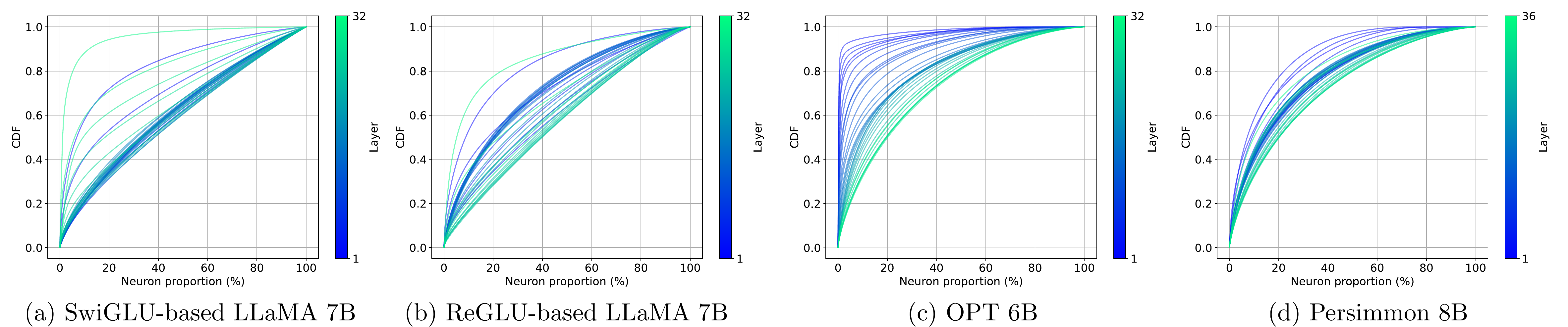}
\caption{Cumulative distribution function (CDF) of neuron activation for different models with about 7B parameters. The x-axis is the neuron proportion (sorted in descending order). OPT 6B uses ReLU as the activation function. Persimmon 8B uses ReLU$^2$ as the activation function.}
\label{fig:commonality_7B}
\end{figure}

Hot-activated neurons refer to the neurons that are activated frequently.
It is beneficial to group these hot-activated neurons together and store them in the cache for better efficiency of parameter access.
Following~\citet{powerinfer}, we calculate the cumulative distribution function (CDF) of neuron activation for different models.
In Figure~\ref{fig:commonality}, we plot the CDF of neuron activation for 1B models.
From this figure, we do not observe a significant existence of hot-activated neurons.
The CDF of neuron activation for variant models with about 7B parameters is shown in Figure~\ref{fig:commonality_7B}.
We can observe a significant existence of hot-activated neurons in some layers of all models.
For example, the top $20\%$ neurons are activated for more than $80\%$ of the time in the lower layers of OPT 6B.
We have not yet fully understood the reasons why 1B models do not show a significant existence of hot-activated neurons while 7B models do.
We want to present the results in the hope of stimulating more research.
One possible reason is that the 1B models are not large enough to exhibit the existence of hot-activated neurons.











\end{document}